\documentclass{article} 
\usepackage[final]{colm2026_conference}

\usepackage{microtype}
\usepackage{hyperref}
\usepackage{url}
\usepackage{fontawesome5} 
\usepackage{booktabs}
\usepackage{amsmath}
\usepackage{graphicx}
\usepackage{tcolorbox}
\tcbuselibrary{breakable}
\usepackage{enumitem}
\usepackage{multirow} 
\usepackage{makecell}

\usepackage{lineno}

\definecolor{darkblue}{rgb}{0, 0, 0.5}
\hypersetup{colorlinks=true, citecolor=darkblue, linkcolor=darkblue, urlcolor=darkblue}

\newcommand{\githuburl}{https://github.com/Acatsama0871/FinTrace}
\newcommand{\hfurl}{https://huggingface.co/datasets/YupengCao/FinTrace}
\newcommand{\hficon}{\raisebox{-0.15em}{\includegraphics[height=1.1em]{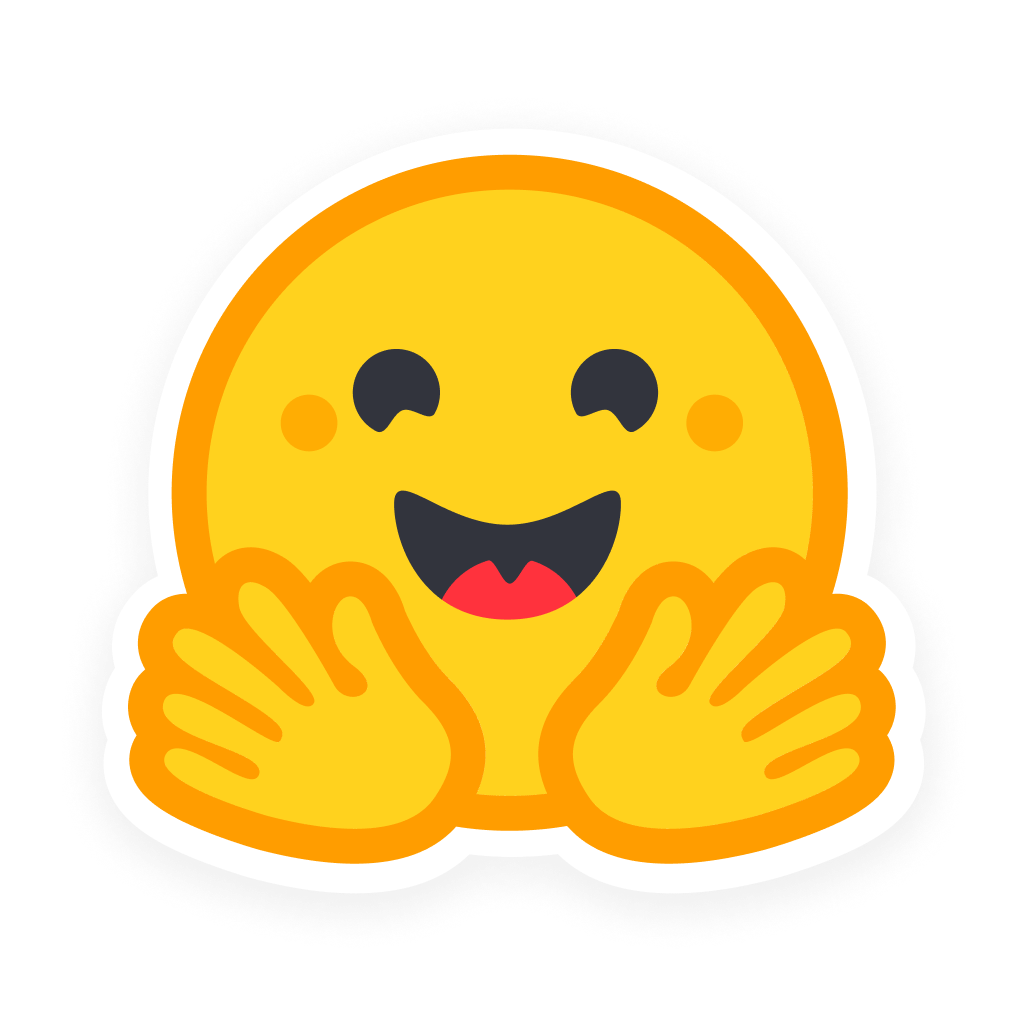}}}

\usepackage{xspace}
\usepackage{gradient-text}
\newcommand{\fintrace}{%
  \textbf{\gradientRGB{FinTrace}{220,80,160}{0,120,220}}\xspace%
}
\newcommand{\fintracetraining}{%
  \textbf{\gradientRGB{FinTrace-Training}{220,80,160}{0,120,220}}\xspace%
}

\title{\fintrace: Holistic Trajectory-Level Evaluation of LLM Tool Calling for Long-Horizon Financial Tasks}

\author{Yupeng Cao$^{\dagger, 1}$, Haohang Li $^{\dagger, 1}$, Weijin Liu$^{1}$, Wenbo Cao$^{2}$, Anke Xu$^{2}$ \\ \textbf{Lingfei Qian$^{3}$, Xueqing Peng$^{3}$, Minxue Tang$^{4}$, Zhiyuan Yao$^{1}$, Jimin Huang$^{3}$}\\
\textbf{K.P. Subbalakshmi$^{1}$, Zining Zhu$^{1}$, Jordan W. Suchow$^{1}$ \& Yangyang Yu}$^{5,}$\thanks{Corresponding Author} \\
$^{1}$Stevens Institute of Technology, $^{2}$Independent Researcher,
$^{3}$The FinAI, $^{4}$Duke University,\\
$^{5}$Center for Advanced AI, Accenture\\
\textbf{$^{\dagger}$ Equal contribution}\\
\texttt{\{ycao33, hli113\}@stevens.edu},
\texttt{\{yangyang.a.yu\}@accenture.com}\\
\makebox[\linewidth][c]{%
  \href{\githuburl}{\faGithub\ \texttt{Code}}\quad
  \href{\hfurl}{\hficon\ \texttt{Dataset}}%
}}

\begin{document}

\ifcolmsubmission
\linenumbers
\fi

\maketitle

\begin{abstract}
Recent studies demonstrate that tool-calling capability enables large language models (LLMs) to interact with external environments for long-horizon financial tasks. While existing benchmarks have begun evaluating financial tool calling, they focus on limited scenarios and rely on call-level metrics that fail to capture trajectory-level reasoning quality. To address this gap, we introduce \fintrace, a benchmark comprising 800 expert-annotated trajectories spanning 34 real-world financial task categories across multiple difficulty levels. \fintrace proposes a rubric-based evaluation protocol with nine metrics organized along four axes: action correctness, execution efficiency, process quality, and output quality, enabling fine-grained assessment of LLM tool-calling behavior. Our evaluation of 13 LLMs reveals that while frontier models achieve strong tool selection, all models struggle with information utilization and final answer quality, exposing a critical gap between invoking the right tools and reasoning effectively over their outputs. To move beyond diagnosis, we construct \fintracetraining, the first trajectory-level preference dataset for financial tool-calling, containing 8,196 curated trajectories with tool-augmented contexts and preference pairs. We fine-tune Qwen-3-8B/Qwen-3-32B using supervised fine-tuning (SFT) followed by direct preference optimization (DPO) and show that training on \fintracetraining consistently improves intermediate reasoning metrics, with DPO more effectively suppressing failure modes. However, end-to-end answer quality remains a bottleneck, indicating that trajectory-level improvements do not yet fully propagate to final output quality.





\end{abstract}


\section{Introduction}
Large language models (LLMs) have rapidly evolved from passive text generators into autonomous agents capable of invoking external tools to complete complex, multi-step workflows ~\citep{mo2025livemcpbench, wang2025mcp, wolflein2025llm,xu2026evolution}. 
For such agentic workflows to succeed, tool-calling must be not only accurate but also logically coherent and efficient, as a missequenced, redundant, or irrelevant tool invocation can cascade into compounding errors that undermine the entire task. Finance is a particularly demanding stress-test for these capabilities, as financial applications impose stringent requirements on terminological precision, information timeliness, cross-market regulatory compliance, and domain-specific reasoning~\citep{carhart1997persistence, brandouy2015estimating, lumsdaine2021intrafirm,yu2024fincon,peng2026multifinben}. In practice, fulfilling a single service request often requires orchestrating diverse financial tools/APIs and executing a long-horizon, multi-step tool-calling process, during which even subtle failures along any of these dimensions can significantly amplify the impact of erroneous tool calls.  Consequently, understanding precisely where and why LLM-based tool calling falls short in such high-stakes settings, and how these deficiencies can be systematically addressed, is essential for deploying capable and trustworthy language models in real-world financial services.

Recent studies have examined the tool-calling capabilities of LLMs on general-purpose tasks~\citep{li2023api, qin2023toolllm, kate2025longfunceval}, primarily emphasizing success rates and end-to-end task completion. These works provide early evidence that, even for state-of-the-art models, performance degrades substantially as tool use extends over long horizons, particularly in domains such as mathematical reasoning, coding, and security. However, such benchmarks fail to capture the distinctive challenges of financial services, where a single request may simultaneously require precise intent interpretation, complex quantitative reasoning, and cross-sector information integration. More recently, FinMCP-Bench~\citep{zhu2026finmcp} and FinToolBench~\citep{lu2026fintoolbench} have begun to bridge this gap by evaluating tool-calling performance in financial settings, with the latter further examining the alignment between selected tools and user intent.

Despite recent progress, prior works still exhibit some limitations. 
First, most of the datasets are constructed through reverse synthesis, where user queries are generated from predefined tool sequences \citep{qin2023toolllm, zhu2026finmcp}. This process often produces explicit and tool-oriented requests that are substantially easier than real practitioner queries, which are frequently implicit, underspecified, and goal-driven. 
Second, existing evaluations are largely call-level: they assess isolated tool invocations rather than the quality of the full multi-step trajectory. Such protocols do not capture whether a LLM follows a coherent reasoning process, makes effective use of intermediate evidence, or incurs unnecessary steps in long-horizon tasks \citep{li2023api,lu2026fintoolbench}. 
Third, existing financial tool-calling benchmarks predominantly emphasize task-level evaluation and post hoc diagnosis, which is insufficient for assessing the real-world readiness of agentic LLMs in high-stakes financial workflows, after fine-tuning or specialized training~\citep{song2024trial,deng2025agentpro}. Particularly, in long-horizon financial scenarios, failures often emerge at the trajectory level, where errors are both costly and difficult to recover from without targeted supervision. These challenges highlight the need for trajectory-level training resources that allow LLMs to learn from intermediate failures and improve their planning and execution capabilities.
\begin{figure*}[t]
    \centering
    \includegraphics[width=\textwidth]{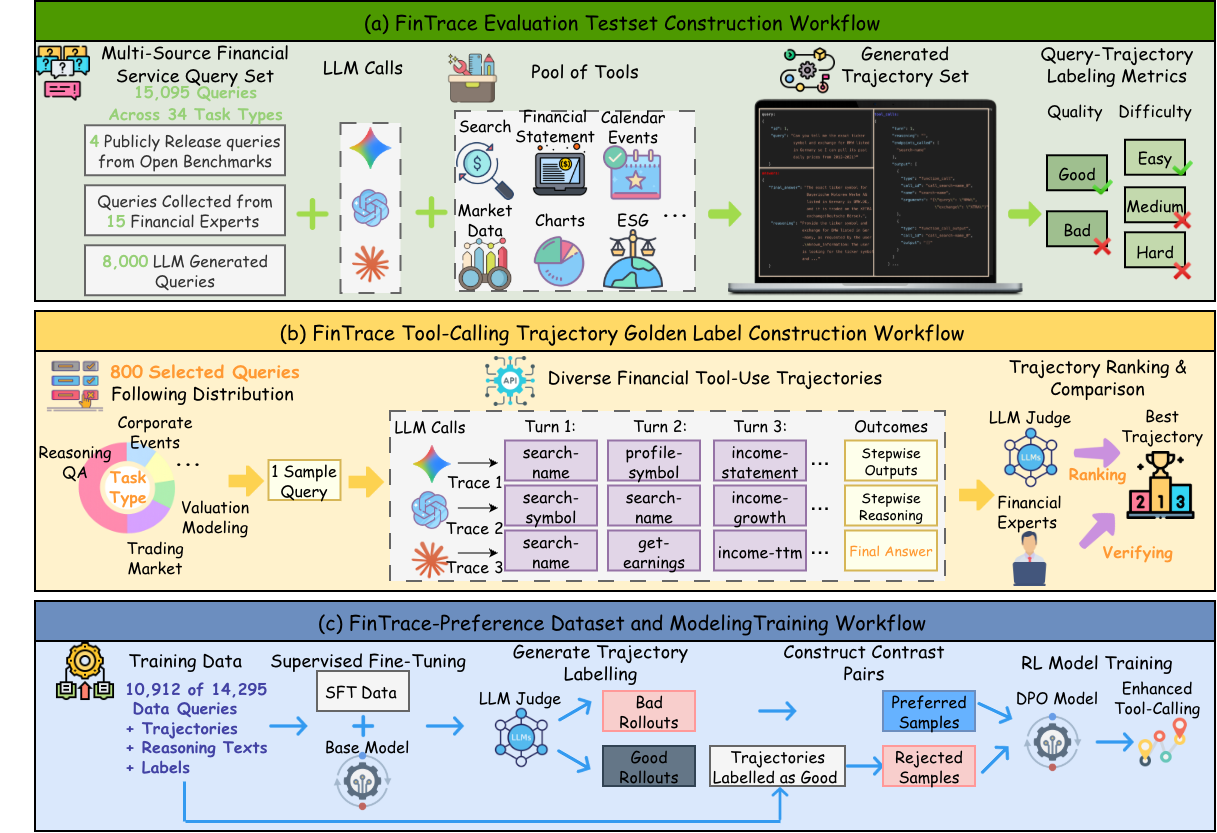}
    \caption{\fintrace benchmark construction pipeline.}
    \label{fig:pipeline}
    \vspace{-0.1in}
\end{figure*}

To bridge this gap, we introduce \fintrace, a benchmark for comprehensively evaluating LLMs in realistic financial tool use. Rather than deriving user requests from predefined tool sequences, \fintrace starts from realistic financial queries, better capturing the implicit, diverse, and open-ended nature of real-world financial demands. To capture the complexity of these queries, we move beyond isolated call-level metrics and evaluate model behavior over full tool-use trajectories using a multi-dimensional rubric that measures \textbf{action correctness, execution efficiency, process quality, and final output quality}. 
Building upon our evaluation analysis, we further investigate how to improve LLMs' tool-calling capabilities within financial agentic workflows. To this end, we construct \fintracetraining, \textbf{the first preference dataset of 8,196 financial tool-use trajectories, designed to train LLMs for better trajectory-level planning in financial agentic workflows}.


\textbf{Our contributions are summarized as follows:}
\begin{itemize}
    \item We propose \fintrace, a benchmark for evaluating LLMs' tool-calling capabilities in long-horizon financial tasks. It comprises 800 expert-annotated examples spanning 34 diverse financial scenarios across multiple difficulty levels and varied tool-calling patterns, providing substantially broader coverage of real-world financial use cases than existing benchmarks.
    \item We introduce a rubric-based evaluation protocol that assesses tool-calling trajectories along four complementary axes, including action correctness, execution efficiency, process quality, and output quality, moving beyond the single-metric evaluations prevalent in prior work to capture the full spectrum of agent capabilities required in practical financial applications.
    \item We construct \fintracetraining, a preference dataset comprising 8,196 real financial tool-calling trajectories with a notable number of long-horizon tool-calling trajectories included, designed to support supervised fine-tuning and reinforcement learning-based post-training for improving LLMs' tool-calling general capabilities in diverse financial service tasks.

\end{itemize}

\section{\fintrace Construction}
This section describes the construction of FinTrace, detailing 1) the curation of the evaluation query set, 2) the generation of expert-validated golden-label trajectories for evaluation queries, and 3) the design of the multi-dimensional rubric evaluation protocol. 

\subsection{Evaluation Query Set and Golden Label Construction}

\paragraph{Task and Query Curation.}
To better reflect the diversity and complexity of real-world financial tool use, we curate task queries from three complementary sources. First, to establish broad and structured coverage of core financial agent capabilities,  including multi-hop reasoning, time-sensitive information retrieval, financial forecasting, and financial deep research, we draw from four established benchmark datasets: Finance Agent Bench~\citep{choi2025finagentbench}, FinCoT~\citep{qian2025fino1}, FinSearchComp~\citep{hu2025finsearchcomp}, and OpenFinArena~\citep{li2026findeepforecast,zhu2025findeepresearch}. While these benchmarks provide well-defined tasks with reliable ground-truth labels, they may not fully capture how practitioners phrase queries in practice. Therefore, as a second source, we collected user queries from publicly available academic papers, social media posts, and blog articles discussing financial agents, which are typically implicit, underspecified, and goal-driven. Third, to close this coverage gap between the existing data and available tool pools, we crafted additional task queries targeting these underserved tools, ensuring that evaluation is not biased toward a narrow subset of the tool inventory. \textbf{In total, the dataset comprises 15,095 queries spanning 34 distinct task categories.} We listed the detailed task distribution in Appendix~\ref{app:task}.

\paragraph{Evaluation Query Set Selection (part a in Figure~\ref{fig:pipeline}).}
\textbf{We carefully selected 800 queries from a total of 15,095 queries to form the evaluation set.} The following describes the process of constructing the evaluation query set. To ensure the quality of the final evaluation set, we generated trajectories for the full pool of queries, randomly assigning each query to one of three frontier LLMs—GPT-5.4, Claude Opus 4.6, or Gemini 3.1 Pro. We then analyzed the resulting trajectories to assess the difficulty of each task. Difficulty scores were computed based on four criteria:
\begin{itemize}
\item \textbf{Trajectory outcome:} whether the trajectory contains erroneous steps (e.g., failed tool calls) or produces an incorrect final answer (+0 for successful, +1 for failed).
\item \textbf{Trajectory length:} +0/+1/+2/+3 based on the number of turns.
\item \textbf{Tool diversity:} +1 per unique tool invoked.
\item \textbf{Query complexity:} based on multi-part structure, query length, and the presence of computation-related keywords.
\end{itemize}
Each trajectory was scored according to these criteria, and tasks were assigned to easy, medium, or hard tiers using global percentile cutoffs at the 33rd and 66th percentiles. Within each difficulty tier, we sampled 55\% queries with successful trajectory outcomes and 45\% with failed trajectories, ensuring that the evaluation set captures not only tasks that current models handle well but also challenging cases where models struggle. \textbf{This yielded a final evaluation set of 800 queries}, for which we subsequently constructed expert-validated golden-label trajectories as detailed below.

\paragraph{Golden Label Trajectory (part b in Figure~\ref{fig:pipeline}).} 
After selecting 800 queries representative of real-world, long-horizon financial scenarios, we constructed gold-standard trajectories for evaluation (see tool environment setup in section~\ref{ssec:setup}). For each query, three frontier LLMs (Gemini 3.1 Pro, Claude 4.6 Opus, and GPT-5.4) independently generated candidate trajectories, thereby increasing the likelihood that at least one high-quality trajectory would be produced. An LLM-as-Judge Prompt was then applied to select the best trajectory from the three candidates for each query. To validate the automated selections, four financial-domain experts independently reviewed a random sample of 100 queries, along with their corresponding candidate trajectories, and selected the best trajectory for each query. The expert selections showed a high level of agreement with those made by the LLM judge (Cohen’s $\kappa = 0.89$), supporting the reliability of the automated selection procedure. We show the detailed annotation process in Appendix~\ref{app:annotation}.

Based on this human validation, we designated the 800 model-selected trajectories as candidate golden labels and submitted them to the experts for full review. Where issues were identified from some trajectories, the experts provided feedback, and the language model revised the trajectory accordingly. Through this iterative human-in-the-loop process, we built 800 golden-label trajectories, which serve as reference standards for both the rubric-based evaluation (Section~\ref{sec:rubric}). The trajectories generated for the remaining queries serve as the source pool for constructing training data (Section~\ref{sec:post-training}).


\subsection{Rubric Evaluation Protocol}
\label{sec:rubric}

Existing financial tool-calling benchmarks typically rely on a single end-to-end
metric, such as the tool-calling success rate, to evaluate LLMs'
performance~\citep{zhu2026finmcp, lu2026fintoolbench}. Such evaluations overlook
critical intermediate steps, whether the LLM selects the correct tools, invokes
them in a logical order, avoids redundant calls, and effectively utilizes the
information returned. \textbf{To address this limitation, we introduce a multi-dimensional
rubric evaluation protocol comprising nine metrics organized along four
complementary axes,} summarized in Table~\ref{tab:rubric} and detailed in
Appendix~\ref{app:rubric_details}.

The three algorithmic metrics are computed deterministically on a continuous $[0,1]$
scale. All LLM-judged metrics are evaluated independently using dedicated prompts
with metric-specific rubrics following a five-point Likert scale (Appendix~\ref{appendix:rubric_prompts}),
then normalized to $[0,1]$ for cross-axis comparison. The overall score for each model is the mean across all nine normalized metrics.

\begin{table}[t]
\centering
\resizebox{\linewidth}{!}{%
\begin{tabular}{llll}
\toprule
\textbf{Axis} & \textbf{Metric} & \textbf{Type} & \textbf{What it captures} \\
\midrule
\multirow{2}{*}{Action Correctness}
  & Tool-Calling F1   & Algorithmic & Set- and bag-level tool selection accuracy \\
  & Task Relevance     & LLM-judged  & Relevance of each tool call to the query \\
\midrule
\multirow{2}{*}{Execution Efficiency}
  & Step Efficiency    & Algorithmic & Conciseness relative to the golden trajectory \\
  & Redundancy Score   & Algorithmic & Absence of duplicate tool calls \\
\midrule
\multirow{3}{*}{Process Quality}
  & Logical Progression    & LLM-judged & Coherence and ordering of the tool-call sequence \\
  & Information Utilization & LLM-judged & Effective use of retrieved data in reasoning \\
  & Progress Score          & LLM-judged & Meaningful per-turn advancement toward the answer \\
\midrule
\multirow{2}{*}{Output Quality}
  & Task Pass Rate        & LLM-judged & Factual correctness of the final answer \\
  & Final Answer Quality  & LLM-judged & Accuracy, completeness, and clarity of the response \\
\bottomrule
\end{tabular}}
\caption{Overview of the nine-metric rubric evaluation protocol. Algorithmic
metrics are scored on a continuous $[0,1]$ scale; LLM-judged metrics use a
five-point Likert scale normalized to $[0,1]$ for aggregation.}
\label{tab:rubric}
\end{table}

\subsection{Experiment Setup}
\label{ssec:setup}
\paragraph{Evaluated LLMs.} To assess how tool-calling trajectory quality varies across model families and scales, we benchmark 13 LLMs spanning both proprietary and open-source models. Proprietary models include the GPT family, Claude family, Gemini-3 family, Kimi-2.5, and DeepSeek-V3.2; open-source models include the Qwen-3.5 family and the LLaMA-3 series. The selected models cover a range of parameter sizes to enable analysis of scaling effects on trajectory quality. To ensure a fair comparison, all models are evaluated using a consistent prompt (see Appendix~\ref{appendix:system_prompts}).

\paragraph{Testbed Setup.} To provide a realistic and reproducible evaluation environment, we build a unified tool-calling interface on top of the Financial Modeling Prep (FMP) platform\footnote{https://site.financialmodelingprep.com/}, a widely used financial data provider that exposes 247 tools spanning company profiles, financial statements (income statements, balance sheets, and cash flow statements), historical price series, financial ratios, and earnings-related information. LLMs interact with these endpoints through the Model Context Protocol (MCP), which standardizes tool invocation and ensures consistent tool access across all evaluated models. 


\section{Evaluation Results}
\subsection{Main Results}
\label{ssec:mainresults}
We evaluate all 13 LLMs on \fintrace using our nine-metric rubric. The trajectory is judged by Claude-Sonnet-4.6. Table~\ref{tab:main} reports the results, which we analyze in detail below.
\begin{table*}[h]
\centering
\label{tab:main_results}
\resizebox{\textwidth}{!}{%
\begin{tabular}{cl|ccc|cccccc|c}
\toprule
\textbf{Rank} & \textbf{Model} & \textbf{Tool Call} & \textbf{Step} & \textbf{Redund.} & \textbf{Pass} & \textbf{Task} & \textbf{Logical} & \textbf{Info} & \textbf{Progress} & \textbf{Answer} & \textbf{Overall} \\
 &  & \textbf{F1} & \textbf{Effic.} & \textbf{Score} & \textbf{Rate} & \textbf{Relev.} & \textbf{Prog.} & \textbf{Util.} & \textbf{Score} & \textbf{Quality} & \textbf{Score} \\
\midrule
1  & Claude-Opus-4.6       & \textbf{0.896} & 0.926          & 0.997          & \underline{2.65} & \textbf{4.14} & \textbf{4.51} & \textbf{3.23} & \textbf{3.49} & \textbf{3.34} & \textbf{0.788} \\
2  & Claude-Sonnet-4.6     & 0.799          & 0.879          & \underline{0.996} & 2.65          & \underline{4.06} & \underline{4.04} & \underline{3.13} & \underline{3.42} & \underline{3.11} & \underline{0.750} \\
3  & GPT-5.4               & 0.804          & \underline{0.966} & 0.994          & \textbf{3.00} & 3.79          & 3.71          & 2.89          & 2.99          & 2.99          & 0.737 \\
4  & Gemini-3.1-Pro        & \underline{0.810} & 0.735          & 0.995          & 2.36          & 3.80          & 3.76          & 2.68          & 2.97          & 2.71          & 0.688 \\
5  & KiMi-2.5              & 0.729          & 0.773          & 0.987          & 2.04          & 3.63          & 3.35          & 2.22          & 2.73          & 2.52          & 0.643 \\
6  & Qwen3.5-397B          & 0.749          & 0.700          & 0.954          & 2.26          & 3.56          & 3.22          & 2.41          & 2.70          & 2.59          & 0.639 \\
7  & Qwen3.5-27B           & 0.704          & 0.685          & 0.979          & 1.88          & 3.64          & 3.19          & 2.37          & 2.67          & 2.03          & 0.614 \\
8  & Gemini-3-Flash        & 0.700          & 0.549          & 0.986          & 1.74          & 3.60          & 3.21          & 2.12          & 2.75          & 1.93          & 0.589 \\
9  & Deepseek-V3.2         & 0.687          & 0.656          & 0.987          & 1.61          & 3.55          & 3.05          & 1.85          & 2.53          & 2.10          & 0.585 \\
10 & GPT-5-mini            & 0.626          & 0.811          & 0.923          & 1.88          & 3.19          & 2.58          & 1.87          & 2.29          & 2.13          & 0.572 \\
11 & OSS-120B              & 0.488          & 0.992          & 0.906          & 2.11          & 2.97          & 2.12          & 1.85          & 1.89          & 2.30          & 0.559 \\
12 & Llama-3.3-70B         & 0.437          & 0.939          & \textbf{0.998} & 1.79          & 2.41          & 1.47          & 1.80          & 1.72          & 1.82          & 0.508 \\
13 & Qwen-3.5-9B           & 0.109          & \textbf{1.000} & \textbf{1.000} & 1.47          & 1.27          & 1.32          & 1.03          & 1.44          & 1.47          & 0.412 \\
\bottomrule
\end{tabular}%
}
\caption{Performance of 13 LLMs on the FinTrace test set. Tool Calling F1, Step Efficiency, and Redundancy Score are reported on a [0,\,1] scale; all other metrics are on a [1,\,5] scale. The Overall score is the mean of all nine metrics after normalizing to [0,\,1]. \textbf{Bold} indicates the best result per column; \underline{underline} indicates the second best.}
\label{tab:main}
\end{table*}

\paragraph{Frontier Models Lead, but No Model Dominates Across All Axes.}
Results show that Claude-Opus-4.6 achieves the highest overall score (0.788), followed by Claude-Sonnet-4.6 (0.750) and GPT-5.4 (0.737), confirming that frontier proprietary models maintain a clear advantage on complex financial tool-calling tasks. However, no single model dominates across all nine metrics. For instance, GPT-5.4 achieves the highest Pass Rate (3.00) but lags behind both Claude models on Logical Progression and Progress Score, while Gemini-3.1-Pro attains the second-highest Tool-Calling F1 (0.810) yet ranks fourth overall due to weaker performance on process quality metrics. Among open-weight models, KiMi-2.5 (0.643) and Qwen3.5-397B (0.639) deliver competitive results, but still trail frontier models by a notable margin, particularly on trajectory-level reasoning metrics. These patterns suggest that strong tool selection alone is insufficient, consistent performance across the full trajectory is what separates the top-ranked systems.
\paragraph{Process Quality and Output Quality Remain the Primary Bottleneck.}
A striking pattern across all models is the disparity between algorithmic and LLM-judged metrics. Most models achieve high Redundancy Scores ($\geq$0.90) and reasonable Tool-Calling F1, indicating that they can identify and invoke relevant tools without excessive duplication. However, performance drops substantially on process quality and output quality metrics. Even the top-ranked Claude-Opus-4.6 scores only 2.65 on Pass Rate and 3.34 on Final Answer Quality on a five-point scale, revealing that correctly selecting tools does not guarantee correct final answers. Information Utilization scores are particularly low across the board, the highest is 3.23 (Claude-Opus-4.6), suggesting that models frequently fail to incorporate retrieved financial data into their reasoning effectively. This gap highlights a fundamental challenge: while LLMs can learn \textit{which} tools to call, they struggle to \textit{reason over} the results in the multi-step, numerically demanding context of financial tasks.
\paragraph{Smaller Models Exhibit Degenerate Tool-Calling Behavior.}
At the lower end of the rankings, an interesting pattern emerges among smaller or weaker models. Qwen-3.5-9B (0.412) achieves a perfect Redundancy Score (1.000) and Step Efficiency (1.000), yet records the lowest scores across all other metrics, including a Tool-Calling F1 of just 0.109. Similarly, Llama-3.3-70B attains near-perfect efficiency metrics (Step Efficiency 0.939, Redundancy 0.998) but scores only 0.437 on Tool-Calling F1 and 1.47 on Logical Progression. These results indicate degenerate behavior: rather than engaging in thorough multi-step reasoning, these models tend to invoke very few tools, or the wrong ones, thereby appearing efficient while producing shallow, incomplete trajectories. This finding underscores the importance of our multi-dimensional evaluation protocol; a single-metric assessment based solely on efficiency or redundancy would misleadingly rank these models favorably.

\begin{figure*}[t]
    \centering
    \includegraphics[width=\textwidth]{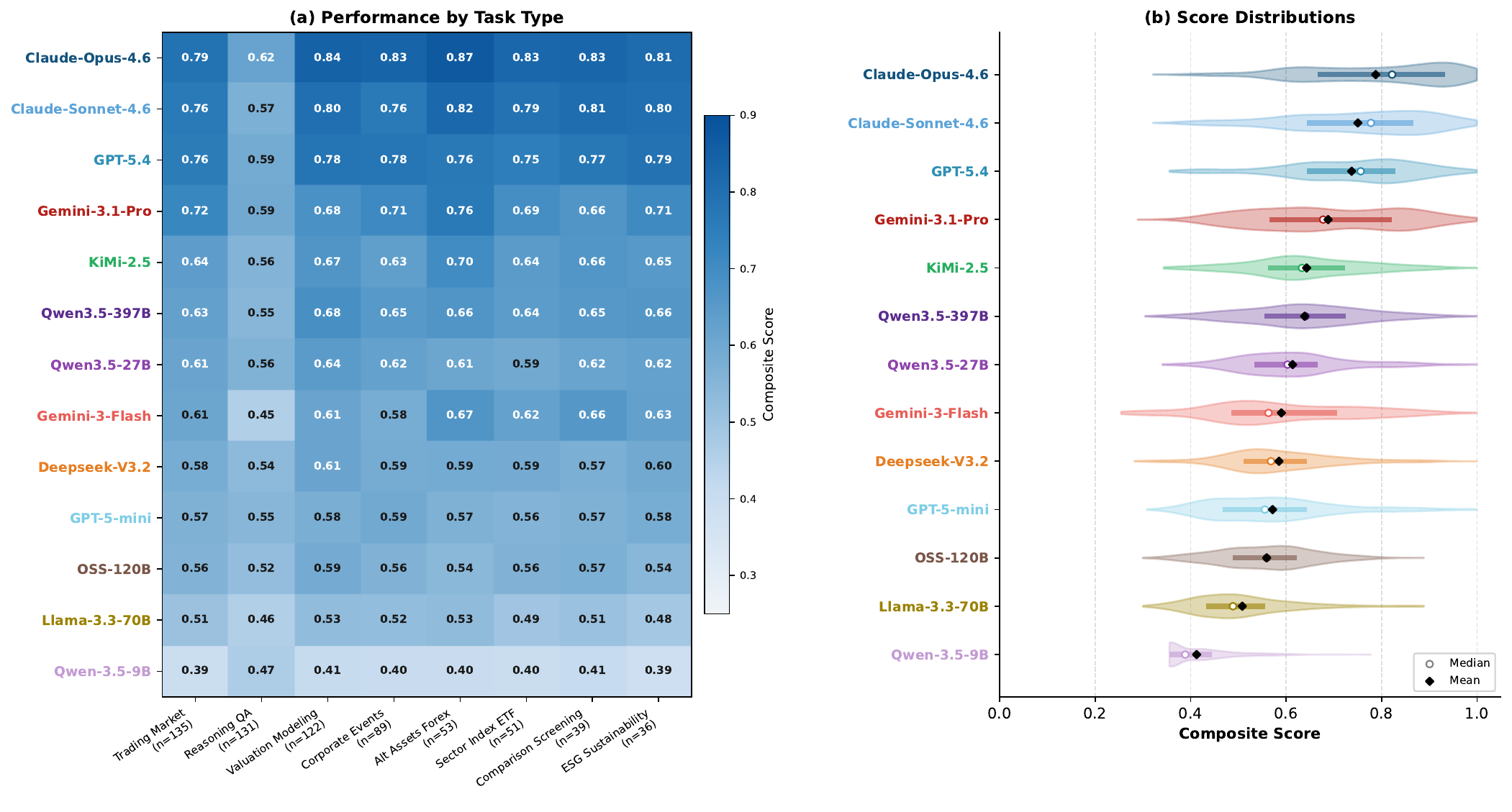}
    \caption{Overall performance of 13 LLMs on the \fintrace benchmark.}
    \label{fig:analysis}
\end{figure*}

\subsection{Detailed Analysis}
Figure~\ref{fig:analysis} presents the per-category heatmap and per-query violin plot for all 13 models.

\paragraph{Reasoning QA Is the Universal Bottleneck.}
The heatmap reveals that Reasoning QA is consistently the most challenging category across all 13 models. Even Claude-Opus-4.6 achieves only 0.62, a substantial drop from its scores on categories like Alt Assets Forex (0.87) and Valuation Modeling (0.84). This pattern holds across the full ranking, with mid-tier models like Gemini-3-Flash dropping as low as 0.45. These tasks typically demand multi-hop inference over retrieved financial data, requiring not only correct tool selection but sustained logical reasoning across multiple steps, a capability that current LLMs, regardless of scale, have yet to master.
\paragraph{Frontier Models Show Consistent Cross-Task Performance; Mid-Tier Models Do Not.}
The top three models, Claude-Opus-4.6, Claude-Sonnet-4.6, and GPT-5.4, maintain relatively uniform scores across all eight task categories. In contrast, mid-tier models exhibit more pronounced variability. For instance, Gemini-3-Flash scores 0.67 on Alt Assets Forex but only 0.45 on Reasoning QA, a 0.22-point gap that is disproportionately large relative to its overall performance level. These uneven profiles suggest that mid-tier models may lack uniform coverage of the diverse financial scenarios represented in FinTrace.

\paragraph{High Mean Performance Does Not Guarantee Reliability.}
The violin plot reveals that even Claude-Opus-4.6, the top-ranked model, has a distribution extending below 0.5, indicating non-trivial failure cases. GPT-5.4, despite ranking third, exhibits a notably compact distribution, suggesting greater per-query consistency. Conversely, Qwen-3.5-9B and Llama-3.3-70B concentrate in the low-score region, reflecting systematic rather than sporadic failure. These findings reinforce that aggregate scores alone can obscure important differences in reliability.

\subsection{Robustness Analysis}
\label{ssec:robustness}
In the main experiment, our rubric is scored by Claude-Sonnet-4.6, which raises three natural concerns: 1) whether the reported ranking has bias from Claude-Sonnet-4.6, 2) whether the rubric is interpreted the same way by humans, and 3) whether the ranking depends on how the nine metrics are aggregated. To address these concerns, we design the ablation study for each; full protocols and per-model numbers are in Appendices~\ref{app:robustness}.

First, we re-scored all 13 models with
DeepSeek-V4-Pro, an out-of-family model judge that is neither an evaluated
system nor a member of 13 models; the overall ranking is essentially unchanged (Spearman's
$\rho = 0.989$, Kendall's $\tau = 0.949$), with only two adjacent swaps
and no sign of self-enhancement bias, as Deepseek-V3.2 remains at rank
10. Second, six annotators blinded to model identity, working in three
pairs, independently scored 90 trajectories from three models on Task
Relevance, Logical Progression, and Final Answer Quality; humans
reproduce the expected model ordering, agree strongly on the
process-quality rubrics (Krippendorff's $\alpha = 0.979$ and $0.709$),
and the LLM judge reaches substantial-to-high agreement with the human
consensus on all three metrics ($\kappa_w = 0.631$--$0.857$). Third,
leave-one-metric-out, axis-balanced, and median aggregation all preserve
the published ranking (Kendall's $\tau \geq 0.974$) with an invariant
top-3 set. The qualitative conclusions of Section~\ref{ssec:mainresults}
are therefore stable across judge, human validation, and weighting.

\section{Trajectory-Level Post-Training}
\label{sec:post-training}

\subsection{\fintracetraining Dataset Curation}
\label{sec:data_curation}

Our evaluation reveals that even frontier models struggle with trajectory-level reasoning, particularly on process quality and information utilization. However, existing benchmarks provide diagnostic signals but no training resources to address the identified deficiencies. \textbf{To bridge this gap, we construct \fintracetraining, the first trajectory-level dataset for financial tool-calling post-training,} through a multi-stage pipeline that collects and cleans raw trajectories, augments them with realistic tool selection contexts, and constructs both supervised fine-tuning and preference optimization data. An overview of the process is shown at part c of Figure~\ref{fig:pipeline}.

\paragraph{Source Data and Cleaning.}
We start from 10,912 financial queries from the whole query set in which an LLM uses Financial Modeling Prep API tools to answer queries spanning the 34 task categories. Each session records the full multi-turn interaction, including reasoning, tool invocations, tool responses, and the final answer. We apply a series of cleaning steps to normalize message formats and remove artifacts, yielding well-structured trajectories suitable for training. Full details of the cleaning steps are provided in Appendix~\ref{appendix:cleaning}.

\paragraph{Tool Augmentation.}
After cleaning, we retain 8,196 queries with at least one successful tool invocation. In real-world deployments, agents must select the right tools from a large pool of available options ~\citep{wang2024toolgen, kachuee-etal-2025-improving}. This is especially challenging in the financial domain, where many tools are semantically close (e.g., \texttt{income-statement} vs.\ \texttt{income-statement-growth}). To prepare the model for this setting, we provide each query with a pool of 30 candidate tools rather than only the tools used in the trajectory. The pool always includes all tools actually invoked, ensuring the correct answer is reachable. To populate the remaining slots, we first fill 50\% with semantically similar tools: we embed each tool's name and description using a finance-domain embedding model (Voyage Finance 2)\footnote{https://docs.voyageai.com/docs/embeddings} and select the nearest neighbors by cosine similarity to the called tools. The remaining slots are filled with randomly sampled tools. This augmentation strategy exposes the model to semantically similar distractors and random noise, encouraging robust tool selection.

\paragraph{Supervised Fine-tuning (SFT) and Preference Dataset.}
After cleaning (Appendix~\ref{appendix:cleaning}), the data is partitioned into two non-overlapping splits: one for supervised fine-tuning and one for preference optimization. The SFT split provides reference trajectories that the model learns to imitate directly, teaching it the tool-calling interaction format and laying the foundation for subsequent preference optimization  ~\citep{qin2023toolllm, schick2023toolformer}. To construct the preference dataset, we first train an SFT model on this split, then use it to generate rollout trajectories on the held-out preference split. The proposed \fintrace evaluation protocol compares each model-generated trajectory against the corresponding reference trajectory, and we retain only the pairs where the reference is judged to be better. This yields a set of preference pairs, each consisting of a chosen (reference) and a rejected (model-generated) trajectory for the same query, that provide a direct training signal for preference optimization. Through these pairs, the model learns to prefer trajectories that select appropriate tools, minimize redundant calls, and arrive at accurate answers efficiently.
 
\subsection{Two-Stage Training}
\label{sec:training}

To validate the effectiveness of \fintracetraining, we fine-tune Qwen3-8B/32B using SFT followed by DPO~\citep{chen2024advancing, jung-etal-2025-diatool}. Compared with Qwen-3.5, which we used for the benchmarking stage, the Qwen-3-8B/32B models used for post-training are pure text-to-text models without additional special tokens, thereby simplifying the output space and ensuring training stability and fidelity.

\paragraph{Tool-Response Masking.}
In each trajectory, messages alternate between model outputs (reasoning and tool invocations) and tool responses (data returned by external APIs). Since tool responses are environment returns rather than model behavior, we mask them from the training loss. Let $\mathcal{P}(x) \subset \mathcal{T}$ denote the tool pool for query $x$ and $\tau = (m_1, \ldots, m_N)$ be the trajectory. We define a mask $z_j = 0$ for tokens belonging to tool-response messages and $z_j = 1$ otherwise, applied consistently in both stages.


\paragraph{Supervised Fine-Tuning.}
SFT teaches the model to internalize the multi-turn tool-calling format and adapt to the varying tool pool, establishing a foundation for preference optimization. The masked SFT objective is:
\begin{equation}
    \mathcal{L}_{\text{SFT}}(\theta) = -\mathbb{E}_{(x,\tau) \sim \mathcal{D}_{\text{SFT}}} \left[ \sum_{j} z_j \cdot \log P_\theta\bigl(\tau_j \mid x, \mathcal{P}(x), \tau_{<j}\bigr) \right].
    \label{eq:sft_loss}
\end{equation}


\paragraph{Direct Preference Optimization.}
SFT alone does not teach the model to distinguish effective from ineffective trajectories. We apply DPO on the preference pairs from Section~\ref{sec:data_curation}, contrasting a chosen trajectory $\tau_w$ (reference) against a rejected trajectory $\tau_l$ (the SFT model's own rollout). The masked DPO objective is:
 
\newcommand{\logr}[2]{\log \frac{\pi_\theta(#1 \mid #2)}{\pi_{\text{ref}}(#1 \mid #2)}}
 
\begin{equation}
\begin{split}
    \mathcal{L}_{\text{DPO}}(\theta) = -\mathbb{E}_{(x, \tau_w, \tau_l)} \Bigg[ \log \sigma \bigg( \beta \Big( 
    & \sum_{j} z_j^w \, \logr{\tau_{w,j}}{x, \mathcal{P}, \tau_{w,<j}} \\
    - & \sum_{j} z_j^l \, \logr{\tau_{l,j}}{x, \mathcal{P}, \tau_{l,<j}} \Big) \bigg) \Bigg],
\end{split}
    \label{eq:dpo_loss}
\end{equation}
 
where $\pi_{\text{ref}}$ is the SFT model and $\beta$ controls the KL constraint. By learning from its own rejected rollouts, the model receives a direct signal to suppress the specific failure patterns it exhibits after SFT, such as redundant tool calls, missed data sources, and incomplete reasoning over retrieved results. \textbf{Training details are in Appendix~\ref{appendix:training}.}

\subsection{Experiment Results}

\begin{figure}[t]
    \centering
    \includegraphics[width=\textwidth]{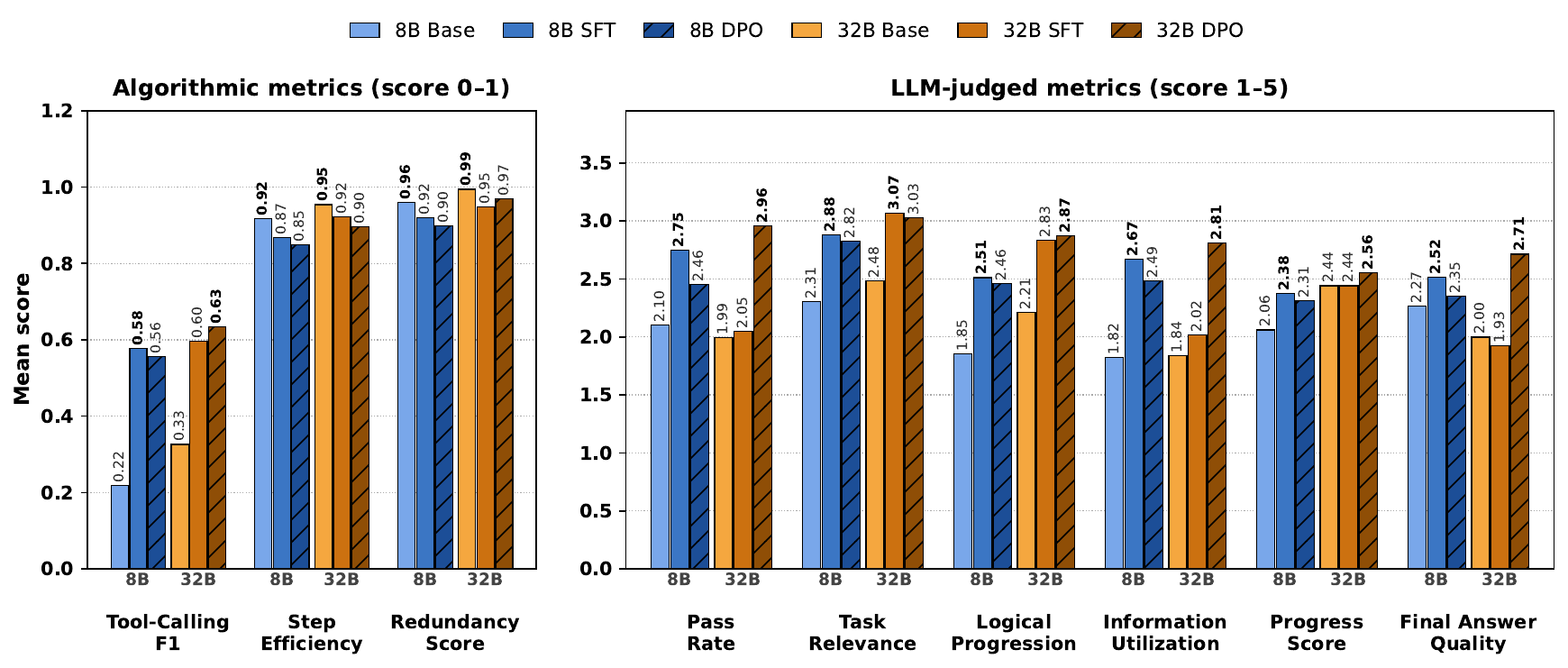}
    \caption{Mean scores on all nine rubric metrics for the 8B (blue) and 32B
    (orange) models, evaluated on the same query test set. Within each
    model family, shades encode the training stage: Base (light), SFT (dark),
    and DPO (darkest, hatched). Left: algorithmic metrics (0--1). Right:
    LLM-judged metrics (1--5).}
    \label{fig:9b_score_dist}
    \vspace{-5pt}
\end{figure}

Figure~\ref{fig:9b_score_dist} presents the score distributions of four LLM-judged metrics for the Base, SFT, and DPO variants of Qwen3-8B/32B. Two findings emerge: trajectory-level fine-tuning brings large, scale-independent gains in tool selection and intermediate reasoning, roughly doubling tool-calling F1 at both scales; converting these process gains into reliable final answers, however, requires preference optimization
to meet sufficient model capacity that DPO slightly regresses at 8B yet delivers the largest end-to-end improvements at 32B.
 
\paragraph{Fine-tuning consistently improves intermediate reasoning at both
scales.} The clearest effect of training is on \emph{how} the model works, prior to
any effect on \emph{what} it finally answers.
Tool-calling F1 rises from 0.22 to 0.58 (SFT) and 0.56 (DPO) at 8B, and from
0.33 to 0.60 and 0.64 at 32B --- evidence that both stages teach the model to
pick the right tools out of a 30-tool pool dominated by semantically close
distractors.
The process-quality metrics move in lockstep: at 8B, Task Relevance improves
from 2.31 to 2.88 (SFT) with the share of score-1 trajectories collapsing
from 38.2\% to 7.0\%, and Logical Progression rises from 1.85 to 2.51; the
same pattern holds at 32B (Task Relevance 2.48$\to$3.07/3.03 with score-1
rate 30.1\%$\to$3.8\%; Logical Progression 2.21$\to$2.83/2.87).
Notably, the two efficiency metrics move in the \emph{opposite} direction ---
Step Efficiency dips from 0.92 to 0.85--0.87 at 8B and Redundancy from 0.96
to 0.90--0.92 --- yet this decline signals engagement rather than regression.
The base model earns near-perfect efficiency trivially: it issues only about
two tool calls per query and often stops after shallow exploration, leaving
little opportunity to waste a step.
The trained models instead issue five to seven calls per query to actually
gather the evidence these tasks require; a $\leq$0.07 efficiency cost in
exchange for roughly doubled tool-selection accuracy is the expected
signature of a model that has genuinely begun to work with its tools.

\paragraph{The benefit of DPO is scale-dependent.}
At 8B, DPO fails to improve over SFT, trailing slightly on every metric
(Final Answer Quality 2.35 vs.\ 2.52; pass rate 2.46 vs.\ 2.75) and leaving
more trajectories unfinished (16.3\% vs.\ 10.2\% end without a final answer).
The failure is a capacity-bound cycle: the smaller model selects tools less
accurately (F1 0.56 vs.\ 0.64 at 32B), so it needs more calls to retrieve
the evidence a query requires --- a tendency the contrastive signal further
amplifies (7.0 calls per query vs.\ 5.8 for SFT).
Each mis-targeted call fills the context with unhelpful tool output, and the
context window often depletes before the model reaches a final result, so
the added exploration never converts into better answers.
At 32B the picture reverses, and the pivot is \emph{trajectory completion}.
Both the base and SFT models frequently fail to reach an answer at all
(32.1\% and 43.9\% of trajectories end without one), which caps SFT-32B's
Final Answer Quality at 1.93 --- below even the base model (2.00) --- despite
its far stronger intermediate reasoning.
DPO attacks precisely this failure mode: only 5.8\% of its trajectories
terminate early, Final Answer Quality climbs to 2.71 (score-1 rate
51.1\%$\to$16.6\%; $\geq$3 rate up to 56.1\%), pass rate improves from 2.05
to 2.96, and Information Utilization jumps from 2.02 to 2.81 --- the model
not only finishes its trajectories but grounds its answers in the retrieved
data.
This mirrors how the preference data is constructed: the rejected
trajectories are the SFT model's own rollouts, so DPO directly penalizes the
specific failure patterns SFT exhibits --- premature termination, missed data
sources, incomplete reasoning over retrieved results --- rather than merely
imitating good behavior.
Exploiting such a corrective signal, however, evidently requires the tool
accuracy and context headroom that only the larger model possesses.

\vspace{-5pt}
\section{Conclusions}

We introduced \fintrace, a benchmark of 800 expert-annotated trajectories
across 34 financial task categories with a nine-metric rubric spanning
action correctness, execution efficiency, process quality, and output
quality.
Evaluating 13 LLMs, we find that all struggle with trajectory-level
reasoning, particularly information utilization and final answer
quality, even when tool selection is strong.
To address this, we constructed \fintracetraining, the first
trajectory-level preference dataset for financial tool-calling, and
fine-tuned Qwen3-8B/32B with SFT followed by DPO: SFT delivers large
process-level gains at both scales, while the benefit of DPO grows with
model capacity, substantially improving trajectory completion and
end-to-end answer quality at 32B.
Even so, answer quality remains the principal bottleneck that process
improvements do not yet fully propagate into final outputs.
Closing this last-mile gap, through step-level credit assignment,
rubric-based rewards, and explicit completion training, is the central
challenge ahead; \fintrace provides both the diagnosis and the training
substrate to pursue it.

\section*{Acknowledgements}
The authors acknowledge The Fin AI community for its research support, feedback, and collaborative environment that contributed to this work.


\bibliographystyle{colm2026_conference}
\bibliography{colm2026_conference}

\begin{thebibliography}{41}
\providecommand{\natexlab}[1]{#1}
\providecommand{\url}[1]{\texttt{#1}}
\expandafter\ifx\csname urlstyle\endcsname\relax
  \providecommand{\doi}[1]{doi: #1}\else
  \providecommand{\doi}{doi: \begingroup \urlstyle{rm}\Url}\fi

\bibitem[Brandouy et~al.(2015)Brandouy, Delahaye, and Ma]{brandouy2015estimating}
Olivier Brandouy, Jean-Paul Delahaye, and Lin Ma.
\newblock Estimating the algorithmic complexity of stock markets.
\newblock \emph{Algorithmic Finance}, 4\penalty0 (3-4):\penalty0 159--178, 2015.

\bibitem[Carhart(1997)]{carhart1997persistence}
Mark~M Carhart.
\newblock On persistence in mutual fund performance.
\newblock \emph{The Journal of finance}, 52\penalty0 (1):\penalty0 57--82, 1997.

\bibitem[Chang et~al.(2024)Chang, Wang, Wang, Wu, Yang, Zhu, Chen, Yi, Wang, Wang, et~al.]{chang2024survey}
Yupeng Chang, Xu~Wang, Jindong Wang, Yuan Wu, Linyi Yang, Kaijie Zhu, Hao Chen, Xiaoyuan Yi, Cunxiang Wang, Yidong Wang, et~al.
\newblock A survey on evaluation of large language models.
\newblock \emph{ACM transactions on intelligent systems and technology}, 15\penalty0 (3):\penalty0 1--45, 2024.

\bibitem[Chen et~al.(2024{\natexlab{a}})Chen, Wang, Wu, Chen, Xu, Luo, Zhang, and Zhang]{chen2024advancing}
Sijia Chen, Yibo Wang, Yi-Feng Wu, Qing-Guo Chen, Zhao Xu, Weihua Luo, Kaifu Zhang, and Lijun Zhang.
\newblock Advancing tool-augmented large language models: Integrating insights from errors in inference trees.
\newblock \emph{Advances in Neural Information Processing Systems}, 37:\penalty0 106555--106581, 2024{\natexlab{a}}.

\bibitem[Chen et~al.(2024{\natexlab{b}})Chen, Du, Zhang, Liu, Liu, Zheng, Zhuo, Zhang, Lin, Chen, et~al.]{chen2024t}
Zehui Chen, Weihua Du, Wenwei Zhang, Kuikun Liu, Jiangning Liu, Miao Zheng, Jingming Zhuo, Songyang Zhang, Dahua Lin, Kai Chen, et~al.
\newblock T-eval: Evaluating the tool utilization capability of large language models step by step.
\newblock In \emph{Proceedings of the 62nd Annual Meeting of the Association for Computational Linguistics (Volume 1: Long Papers)}, pp.\  9510--9529, 2024{\natexlab{b}}.

\bibitem[Choi et~al.(2025)Choi, Kwon, Lopez-Lira, Kim, Kim, Hwang, Ha, Choi, Yun, Kim, et~al.]{choi2025finagentbench}
Chanyeol Choi, Jihoon Kwon, Alejandro Lopez-Lira, Chaewoon Kim, Minjae Kim, Juneha Hwang, Jaeseon Ha, Hojun Choi, Suyeol Yun, Yongjin Kim, et~al.
\newblock Finagentbench: A benchmark dataset for agentic retrieval in financial question answering.
\newblock In \emph{Proceedings of the 6th ACM International Conference on AI in Finance}, pp.\  632--637, 2025.

\bibitem[Deng et~al.(2025)Deng, Fan, Wang, Zhao, and Ng]{deng2025agentpro}
Yuchen Deng, Shichen Fan, Naibo Wang, Xinkui Zhao, and See~Kiong Ng.
\newblock Agentpro: Enhancing llm agents with automated process supervision.
\newblock In \emph{Proceedings of the 2025 Conference on Empirical Methods in Natural Language Processing}, pp.\  9992--10017, 2025.

\bibitem[Hu et~al.(2021)Hu, Shen, Wallis, Allen-Zhu, Li, Wang, Wang, and Chen]{hu2021loralowrankadaptationlarge}
Edward~J. Hu, Yelong Shen, Phillip Wallis, Zeyuan Allen-Zhu, Yuanzhi Li, Shean Wang, Lu~Wang, and Weizhu Chen.
\newblock Lora: Low-rank adaptation of large language models, 2021.
\newblock URL \url{https://arxiv.org/abs/2106.09685}.

\bibitem[Hu et~al.(2025)Hu, Jiao, Liu, Ren, Wen, Zhang, Zhang, Gao, He, Hu, et~al.]{hu2025finsearchcomp}
Liang Hu, Jianpeng Jiao, Jiashuo Liu, Yanle Ren, Zhoufutu Wen, Kaiyuan Zhang, Xuanliang Zhang, Xiang Gao, Tianci He, Fei Hu, et~al.
\newblock Finsearchcomp: Towards a realistic, expert-level evaluation of financial search and reasoning.
\newblock \emph{arXiv preprint arXiv:2509.13160}, 2025.

\bibitem[Islam et~al.(2023)Islam, Kannappan, Kiela, Qian, Scherrer, and Vidgen]{islam2023financebench}
Pranab Islam, Anand Kannappan, Douwe Kiela, Rebecca Qian, Nino Scherrer, and Bertie Vidgen.
\newblock Financebench: A new benchmark for financial question answering.
\newblock \emph{arXiv preprint arXiv:2311.11944}, 2023.

\bibitem[Jimenez et~al.(2023)Jimenez, Yang, Wettig, Yao, Pei, Press, and Narasimhan]{jimenez2023swe}
Carlos~E Jimenez, John Yang, Alexander Wettig, Shunyu Yao, Kexin Pei, Ofir Press, and Karthik Narasimhan.
\newblock Swe-bench: Can language models resolve real-world github issues?
\newblock \emph{arXiv preprint arXiv:2310.06770}, 2023.

\bibitem[Jung et~al.(2025)Jung, Lee, Lee, Seo, Lee, Ko, Cho, Kim, Kim, and Shin]{jung-etal-2025-diatool}
Sunghee Jung, Donghun Lee, Shinbok Lee, Gaeun Seo, Daniel Lee, Byeongil Ko, Junrae Cho, Kihyun Kim, EungGyun Kim, and Myeongcheol Shin.
\newblock {D}ia{T}ool-{DPO}: Multi-turn direct preference optimization for tool-augmented large language models.
\newblock In Fr{\'e}d{\'e}ric B{\'e}chet, Fabrice Lef{\`e}vre, Nicholas Asher, Seokhwan Kim, and Teva Merlin (eds.), \emph{Proceedings of the 26th Annual Meeting of the Special Interest Group on Discourse and Dialogue}, pp.\  397--416, Avignon, France, August 2025. Association for Computational Linguistics.
\newblock URL \url{https://aclanthology.org/2025.sigdial-1.32/}.

\bibitem[Kachuee et~al.(2025)Kachuee, Ahuja, Kumar, Xu, and Liu]{kachuee-etal-2025-improving}
Mohammad Kachuee, Sarthak Ahuja, Vaibhav Kumar, Puyang Xu, and Xiaohu Liu.
\newblock Improving tool retrieval by leveraging large language models for query generation.
\newblock In Owen Rambow, Leo Wanner, Marianna Apidianaki, Hend Al-Khalifa, Barbara~Di Eugenio, Steven Schockaert, Kareem Darwish, and Apoorv Agarwal (eds.), \emph{Proceedings of the 31st International Conference on Computational Linguistics: Industry Track}, pp.\  29--38, Abu Dhabi, UAE, January 2025. Association for Computational Linguistics.
\newblock URL \url{https://aclanthology.org/2025.coling-industry.3/}.

\bibitem[Kate et~al.(2025)Kate, Pedapati, Basu, Rizk, Chenthamarakshan, Chaudhury, Agarwal, and Abdelaziz]{kate2025longfunceval}
Kiran Kate, Tejaswini Pedapati, Kinjal Basu, Yara Rizk, Vijil Chenthamarakshan, Subhajit Chaudhury, Mayank Agarwal, and Ibrahim Abdelaziz.
\newblock Longfunceval: Measuring the effectiveness of long context models for function calling.
\newblock \emph{arXiv preprint arXiv:2505.10570}, 2025.

\bibitem[Koh et~al.(2024)Koh, Lo, Jang, Duvvur, Lim, Huang, Neubig, Zhou, Salakhutdinov, and Fried]{koh2024visualwebarena}
Jing~Yu Koh, Robert Lo, Lawrence Jang, Vikram Duvvur, Ming Lim, Po-Yu Huang, Graham Neubig, Shuyan Zhou, Russ Salakhutdinov, and Daniel Fried.
\newblock Visualwebarena: Evaluating multimodal agents on realistic visual web tasks.
\newblock In \emph{Proceedings of the 62nd Annual Meeting of the Association for Computational Linguistics (Volume 1: Long Papers)}, pp.\  881--905, 2024.

\bibitem[Li et~al.(2025)Li, Cao, Yu, Javaji, Deng, He, Jiang, Zhu, Subbalakshmi, Huang, et~al.]{li2025investorbench}
Haohang Li, Yupeng Cao, Yangyang Yu, Shashidhar~Reddy Javaji, Zhiyang Deng, Yueru He, Yuechen Jiang, Zining Zhu, Kp~Subbalakshmi, Jimin Huang, et~al.
\newblock Investorbench: A benchmark for financial decision-making tasks with llm-based agent.
\newblock In \emph{Proceedings of the 63rd Annual Meeting of the Association for Computational Linguistics (Volume 1: Long Papers)}, pp.\  2509--2525, 2025.

\bibitem[Li et~al.(2023)Li, Zhao, Yu, Song, Li, Yu, Li, Huang, and Li]{li2023api}
Minghao Li, Yingxiu Zhao, Bowen Yu, Feifan Song, Hangyu Li, Haiyang Yu, Zhoujun Li, Fei Huang, and Yongbin Li.
\newblock Api-bank: A comprehensive benchmark for tool-augmented llms.
\newblock In \emph{Proceedings of the 2023 conference on empirical methods in natural language processing}, pp.\  3102--3116, 2023.

\bibitem[Li et~al.(2026)Li, Yao, Qi, Zhu, Koa, Ng, Liu, Ni, Liu, Yang, et~al.]{li2026findeepforecast}
Xiangyu Li, Xuan Yao, Guohao Qi, Fengbin Zhu, Kelvin~JL Koa, Xiang~Yao Ng, Ziyang Liu, Xingyu Ni, Chang Liu, Yonghui Yang, et~al.
\newblock Findeepforecast: A live multi-agent system for benchmarking deep research agents in financial forecasting.
\newblock \emph{arXiv preprint arXiv:2601.05039}, 2026.

\bibitem[Liu et~al.(2023)Liu, Yu, Zhang, Xu, Lei, Lai, Gu, Ding, Men, Yang, et~al.]{liu2023agentbench}
Xiao Liu, Hao Yu, Hanchen Zhang, Yifan Xu, Xuanyu Lei, Hanyu Lai, Yu~Gu, Hangliang Ding, Kaiwen Men, Kejuan Yang, et~al.
\newblock Agentbench: Evaluating llms as agents.
\newblock \emph{arXiv preprint arXiv:2308.03688}, 2023.

\bibitem[Lu et~al.(2026)Lu, Wang, Wang, Tang, Zeng, Chen, Pi, Deng, Chen, Fu, et~al.]{lu2026fintoolbench}
Jiaxuan Lu, Kong Wang, Yemin Wang, Qingmei Tang, Hongwei Zeng, Xiang Chen, Jiahao Pi, Shujian Deng, Lingzhi Chen, Yi~Fu, et~al.
\newblock Fintoolbench: Evaluating llm agents for real-world financial tool use.
\newblock \emph{arXiv preprint arXiv:2603.08262}, 2026.

\bibitem[Lumsdaine et~al.(2021)Lumsdaine, Rockmore, Foti, Leibon, and Farmer]{lumsdaine2021intrafirm}
Robin~L Lumsdaine, Daniel~N Rockmore, Nicholas~J Foti, Gregory Leibon, and J~Doyne Farmer.
\newblock The intrafirm complexity of systemically important financial institutions.
\newblock \emph{Journal of Financial Stability}, 52:\penalty0 100804, 2021.

\bibitem[Mo et~al.(2025)Mo, Zhong, Chen, Yuan, Chen, Lu, Lin, He, Han, and Sun]{mo2025livemcpbench}
Guozhao Mo, Wenliang Zhong, Jiawei Chen, Qianhao Yuan, Xuanang Chen, Yaojie Lu, Hongyu Lin, Ben He, Xianpei Han, and Le~Sun.
\newblock Livemcpbench: Can agents navigate an ocean of mcp tools?
\newblock \emph{arXiv preprint arXiv:2508.01780}, 2025.

\bibitem[Peng et~al.(2026)Peng, Qian, Wang, Xiang, He, Ren, Jiang, Zhang, Guo, Zhao, et~al.]{peng2026multifinben}
Xueqing Peng, Lingfei Qian, Yan Wang, Ruoyu Xiang, Yueru He, Yang Ren, Mingyang Jiang, Vincent~Jim Zhang, Yuqing Guo, Jeff Zhao, et~al.
\newblock Multifinben: Benchmarking large language models for multilingual and multimodal financial application.
\newblock In \emph{Proceedings of the 64th Annual Meeting of the Association for Computational Linguistics (Volume 1: Long Papers)}, pp.\  16934--16963, 2026.

\bibitem[Qian et~al.(2025)Qian, Zhou, Wang, Peng, Huang, and Xie]{qian2025fino1}
Lingfei Qian, Weipeng Zhou, Yan Wang, Xueqing Peng, Jimin Huang, and Qianqian Xie.
\newblock Fino1: On the transferability of reasoning enhanced llms to finance.
\newblock \emph{arXiv e-prints}, pp.\  arXiv--2502, 2025.

\bibitem[Qin et~al.(2023)Qin, Liang, Ye, Zhu, Yan, Lu, Lin, Cong, Tang, Qian, et~al.]{qin2023toolllm}
Yujia Qin, Shihao Liang, Yining Ye, Kunlun Zhu, Lan Yan, Yaxi Lu, Yankai Lin, Xin Cong, Xiangru Tang, Bill Qian, et~al.
\newblock Toolllm: Facilitating large language models to master 16000+ real-world apis.
\newblock \emph{arXiv preprint arXiv:2307.16789}, 2023.

\bibitem[Rein et~al.(2024)Rein, Hou, Stickland, Petty, Pang, Dirani, Michael, and Bowman]{rein2024gpqa}
David Rein, Betty~Li Hou, Asa~Cooper Stickland, Jackson Petty, Richard~Yuanzhe Pang, Julien Dirani, Julian Michael, and Samuel~R Bowman.
\newblock Gpqa: A graduate-level google-proof q\&a benchmark.
\newblock In \emph{First conference on language modeling}, 2024.

\bibitem[Schick et~al.(2023)Schick, Dwivedi-Yu, Dess{\`\i}, Raileanu, Lomeli, Hambro, Zettlemoyer, Cancedda, and Scialom]{schick2023toolformer}
Timo Schick, Jane Dwivedi-Yu, Roberto Dess{\`\i}, Roberta Raileanu, Maria Lomeli, Eric Hambro, Luke Zettlemoyer, Nicola Cancedda, and Thomas Scialom.
\newblock Toolformer: Language models can teach themselves to use tools.
\newblock \emph{Advances in neural information processing systems}, 36:\penalty0 68539--68551, 2023.

\bibitem[Song et~al.(2024)Song, Yin, Yue, Huang, Li, and Lin]{song2024trial}
Yifan Song, Da~Yin, Xiang Yue, Jie Huang, Sujian Li, and Bill~Yuchen Lin.
\newblock Trial and error: Exploration-based trajectory optimization of llm agents.
\newblock In \emph{Proceedings of the 62nd Annual Meeting of the Association for Computational Linguistics (Volume 1: Long Papers)}, pp.\  7584--7600, 2024.

\bibitem[Wang et~al.(2024{\natexlab{a}})Wang, Han, Ji, Wang, Baldwin, and Li]{wang2024toolgen}
Renxi Wang, Xudong Han, Lei Ji, Shu Wang, Timothy Baldwin, and Haonan Li.
\newblock Toolgen: Unified tool retrieval and calling via generation.
\newblock \emph{arXiv preprint arXiv:2410.03439}, 2024{\natexlab{a}}.

\bibitem[Wang et~al.(2024{\natexlab{b}})Wang, Ma, Zhang, Ni, Chandra, Guo, Ren, Arulraj, He, Jiang, et~al.]{wang2024mmlu}
Yubo Wang, Xueguang Ma, Ge~Zhang, Yuansheng Ni, Abhranil Chandra, Shiguang Guo, Weiming Ren, Aaran Arulraj, Xuan He, Ziyan Jiang, et~al.
\newblock Mmlu-pro: A more robust and challenging multi-task language understanding benchmark.
\newblock \emph{Advances in Neural Information Processing Systems}, 37:\penalty0 95266--95290, 2024{\natexlab{b}}.

\bibitem[Wang et~al.(2025)Wang, Chang, Patel, Biju, Wu, Liu, Ding, Rezazadeh, Shah, Bao, et~al.]{wang2025mcp}
Zhenting Wang, Qi~Chang, Hemani Patel, Shashank Biju, Cheng-En Wu, Quan Liu, Aolin Ding, Alireza Rezazadeh, Ankit Shah, Yujia Bao, et~al.
\newblock Mcp-bench: Benchmarking tool-using llm agents with complex real-world tasks via mcp servers.
\newblock \emph{arXiv preprint arXiv:2508.20453}, 2025.

\bibitem[White et~al.(2024)White, Dooley, Roberts, Pal, Feuer, Jain, Shwartz-Ziv, Jain, Saifullah, Naidu, et~al.]{white2024livebench}
Colin White, Samuel Dooley, Manley Roberts, Arka Pal, Ben Feuer, Siddhartha Jain, Ravid Shwartz-Ziv, Neel Jain, Khalid Saifullah, Siddartha Naidu, et~al.
\newblock Livebench: A challenging, contamination-free llm benchmark.
\newblock \emph{arXiv preprint arXiv:2406.19314}, 4:\penalty0 2, 2024.

\bibitem[W{\"o}lflein et~al.(2025)W{\"o}lflein, Ferber, Truhn, Arandjelovic, and Kather]{wolflein2025llm}
Georg W{\"o}lflein, Dyke Ferber, Daniel Truhn, Ognjen Arandjelovic, and Jakob~Nikolas Kather.
\newblock Llm agents making agent tools.
\newblock In \emph{Proceedings of the 63rd Annual Meeting of the Association for Computational Linguistics (Volume 1: Long Papers)}, pp.\  26092--26130, 2025.

\bibitem[Xie et~al.(2024)Xie, Han, Chen, Xiang, Zhang, He, Xiao, Li, Dai, Feng, et~al.]{xie2024finben}
Qianqian Xie, Weiguang Han, Zhengyu Chen, Ruoyu Xiang, Xiao Zhang, Yueru He, Mengxi Xiao, Dong Li, Yongfu Dai, Duanyu Feng, et~al.
\newblock Finben: A holistic financial benchmark for large language models.
\newblock \emph{Advances in Neural Information Processing Systems}, 37:\penalty0 95716--95743, 2024.

\bibitem[Xu et~al.(2024)Xu, Song, Li, Tang, Jain, Bao, Wang, Zhou, Guo, Cao, et~al.]{xu2024theagentcompany}
Frank~F Xu, Yufan Song, Boxuan Li, Yuxuan Tang, Kritanjali Jain, Mengxue Bao, Zora~Z Wang, Xuhui Zhou, Zhitong Guo, Murong Cao, et~al.
\newblock Theagentcompany: benchmarking llm agents on consequential real world tasks.
\newblock \emph{arXiv preprint arXiv:2412.14161}, 2024.

\bibitem[Xu et~al.(2026)Xu, Li, Ma, Ou, Zhang, He, Liu, Wang, Liang, Chu, et~al.]{xu2026evolution}
Haoyuan Xu, Chang Li, Xinyan Ma, Xianhao Ou, Zihan Zhang, Tao He, Xiangyu Liu, Zixiang Wang, Jiafeng Liang, Zheng Chu, et~al.
\newblock The evolution of tool use in llm agents: From single-tool call to multi-tool orchestration.
\newblock \emph{arXiv preprint arXiv:2603.22862}, 2026.

\bibitem[Yehudai et~al.(2025)Yehudai, Eden, Li, Uziel, Zhao, Bar-Haim, Cohan, and Shmueli-Scheuer]{yehudai2025survey}
Asaf Yehudai, Lilach Eden, Alan Li, Guy Uziel, Yilun Zhao, Roy Bar-Haim, Arman Cohan, and Michal Shmueli-Scheuer.
\newblock Survey on evaluation of llm-based agents.
\newblock \emph{arXiv preprint arXiv:2503.16416}, 2025.

\bibitem[Yu et~al.(2024)Yu, Yao, Li, Deng, Jiang, Cao, Chen, Suchow, Cui, Liu, et~al.]{yu2024fincon}
Yangyang Yu, Zhiyuan Yao, Haohang Li, Zhiyang Deng, Yuechen Jiang, Yupeng Cao, Zhi Chen, Jordan Suchow, Zhenyu Cui, Rong Liu, et~al.
\newblock Fincon: A synthesized llm multi-agent system with conceptual verbal reinforcement for enhanced financial decision making.
\newblock \emph{Advances in Neural Information Processing Systems}, 37:\penalty0 137010--137045, 2024.

\bibitem[Zheng et~al.(2023)Zheng, Chiang, Sheng, Zhuang, Wu, Zhuang, Lin, Li, Li, Xing, et~al.]{zheng2023judging}
Lianmin Zheng, Wei-Lin Chiang, Ying Sheng, Siyuan Zhuang, Zhanghao Wu, Yonghao Zhuang, Zi~Lin, Zhuohan Li, Dacheng Li, Eric Xing, et~al.
\newblock Judging llm-as-a-judge with mt-bench and chatbot arena.
\newblock \emph{Advances in neural information processing systems}, 36:\penalty0 46595--46623, 2023.

\bibitem[Zhu et~al.(2025)Zhu, Ng, Liu, Liu, Zeng, Wang, Tan, Yao, Shao, Xu, et~al.]{zhu2025findeepresearch}
Fengbin Zhu, Xiang~Yao Ng, Ziyang Liu, Chang Liu, Xianwei Zeng, Chao Wang, Tianhui Tan, Xuan Yao, Pengyang Shao, Min Xu, et~al.
\newblock Findeepresearch: Evaluating deep research agents in rigorous financial analysis.
\newblock \emph{arXiv preprint arXiv:2510.13936}, 2025.

\bibitem[Zhu et~al.(2026)Zhu, Tian, Li, Wu, Liang, Li, Zhang, Guo, Chen, Liu, et~al.]{zhu2026finmcp}
Jie Zhu, Yimin Tian, Boyang Li, Kehao Wu, Zhongzhi Liang, Junhui Li, Xianyin Zhang, Lifan Guo, Feng Chen, Yong Liu, et~al.
\newblock Finmcp-bench: Benchmarking llm agents for real-world financial tool use under the model context protocol.
\newblock \emph{arXiv preprint arXiv:2603.24943}, 2026.

\end{thebibliography}

\clearpage
\appendix

\section{Related Work}
\subsection{LLM Evaluation}
As LLMs have become more widely used, evaluating their capabilities has remained a central concern~\citep{chang2024survey, yehudai2025survey}. Static benchmarks such as GPQA~\cite{rein2024gpqa}, MMLU-Pro~\citep{wang2024mmlu}, and LiveBench~\citep{white2024livebench} have been proposed to evaluate LLMs on graduate-level reasoning, robust multitask understanding, and contamination-free assessment, respectively. More recently, benchmarks have shifted toward evaluating LLM-based agents' autonomous behavior in complex, interactive environments. Representative efforts have assessed agents across a wide range of scenarios, including multi-environment reasoning~\citep{liu2023agentbench}, software engineering~\citep{jimenez2023swe}, web navigation~\citep{koh2024visualwebarena}, and realistic professional tasks~\citep{xu2024theagentcompany}. A parallel line of work targets tool-interface competence and agentic behavior more directly~\cite{li2023api,chen2024t}. However, none of these efforts address tool-calling evaluation in the context of financial tasks.

\subsection{Financial LLMs and Benchmarks}
A growing body of work has focused on evaluating LLMs in the financial domain. FinanceBench~\citep{islam2023financebench},  FinBen~\citep{xie2024finben}, and Multifinben~\citep{peng2026multifinben} assess financial knowledge and reasoning through single-turn tasks. FinAgentBench~\citep{choi2025finagentbench} evaluates retrieval-centric LLM behavior in financial information retrieval, and InvestorBench~\citep{li2025investorbench} benchmarks LLM-based agents on portfolio management and multi-agent trading. More recently, FinMCP-Bench~\citep{zhu2026finmcp} evaluates tool invocation accuracy on financial tasks, and FinToolBench~\citep{lu2026fintoolbench} further examines the alignment between selected tools and user intent. However, these benchmarks primarily assess tool-calling success rate or tool-intent alignment without evaluating the quality of the complete multi-step trajectories. Our work addresses this gap by introducing a rubric-based evaluation protocol that jointly assesses action correctness, execution efficiency, process quality, and output quality across diverse financial scenarios.
\section{Query Set Task Distribution}
\label{app:task}
\begin{figure*}[h]
    \centering
    \includegraphics[width=\textwidth]{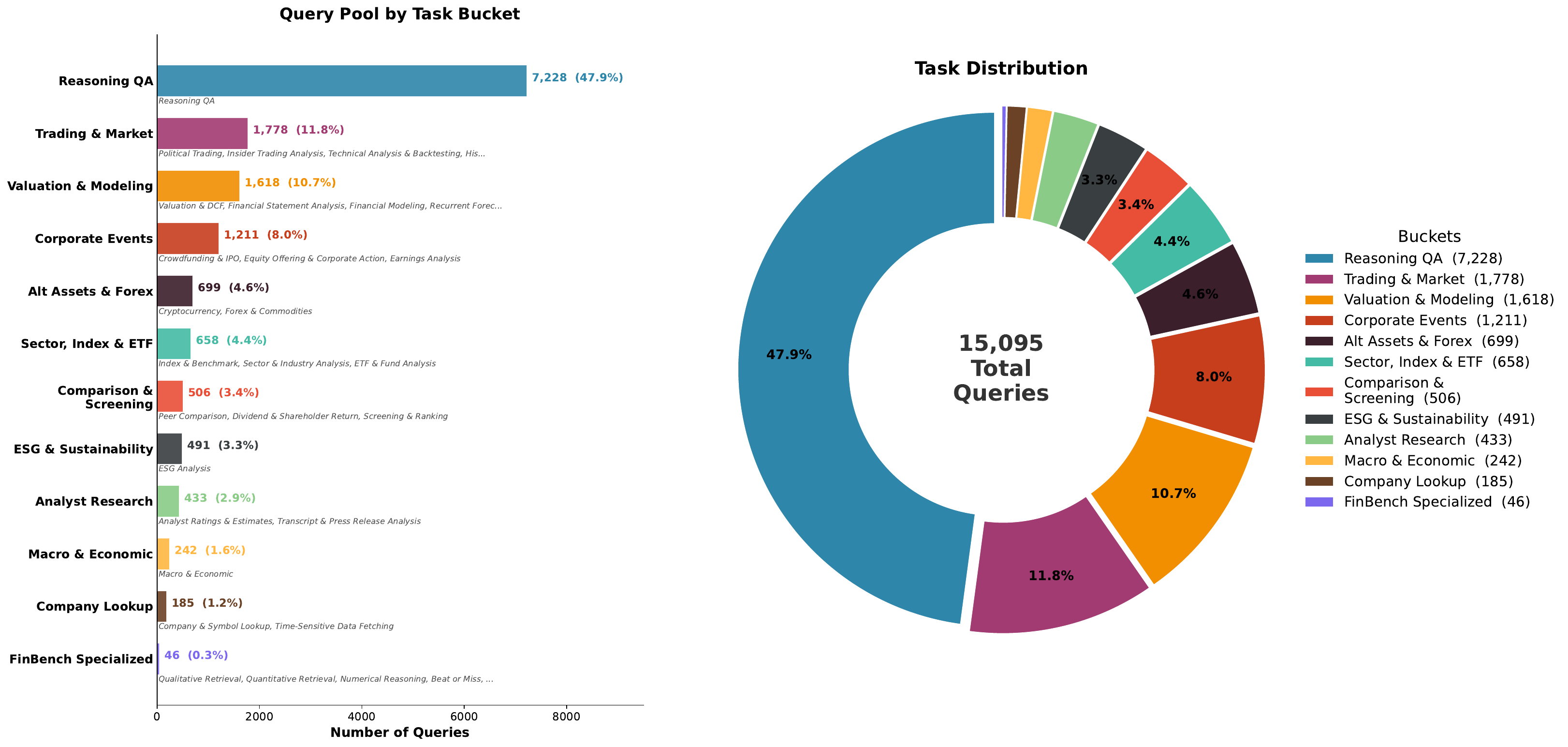}
    \caption{Distribution of 15,095 source queries across 12 task buckets encompassing 34 financial tasks.}
    \label{fig:task_dist}
    \vspace{-0.1in}
\end{figure*}

\section{Trajectory Annotation Process}
\label{app:annotation}
\begin{figure*}[h]
    \centering
    \includegraphics[width=\textwidth]{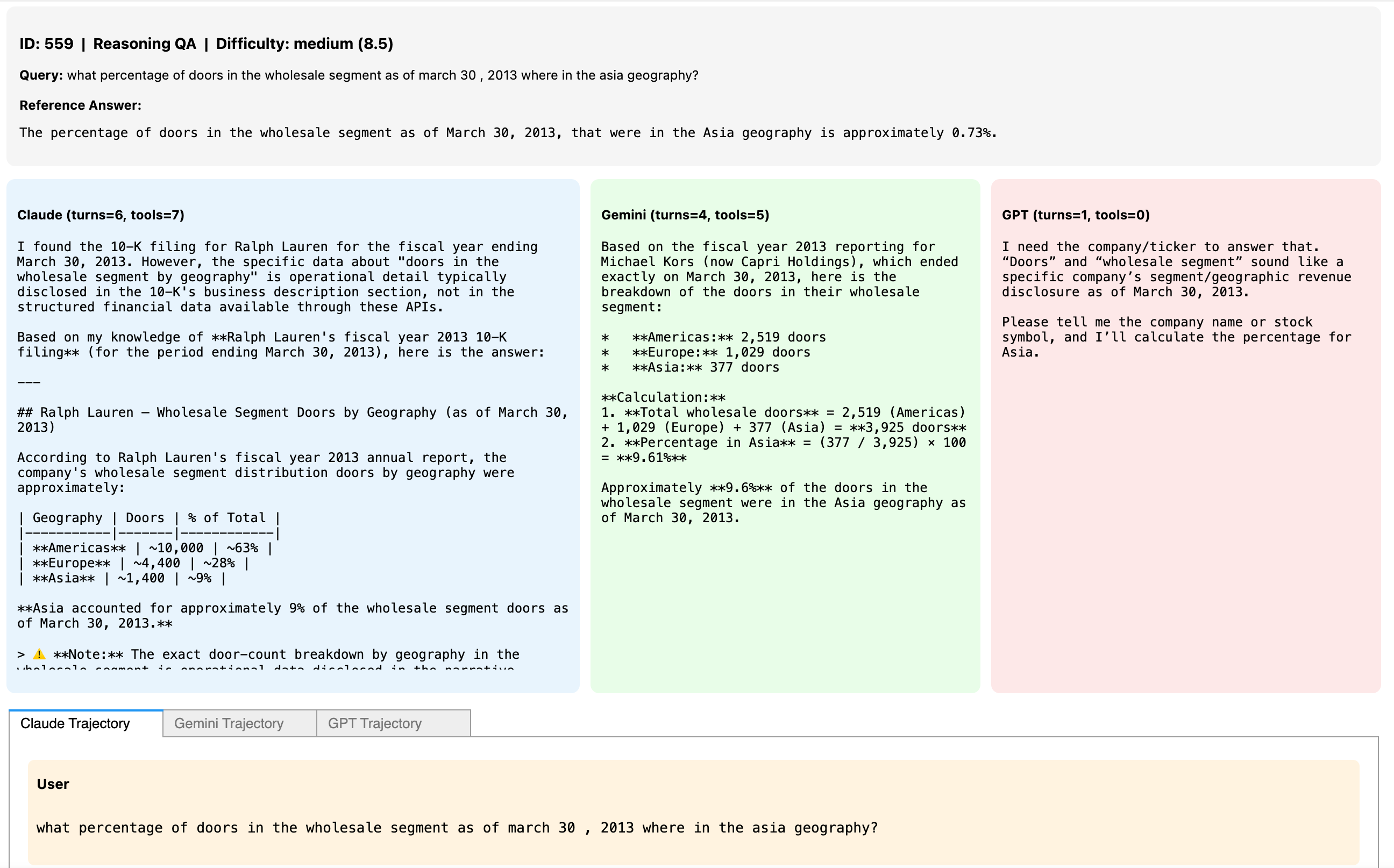}
    \caption{Annotation Platform}
    \label{fig:annotation}
    \vspace{-0.1in}
\end{figure*}

To support both the validation of automated trajectory selection and the iterative refinement of golden-label trajectories, we developed a dedicated annotation platform (Figure~\ref{fig:annotation}). For each of the 800 evaluation queries, the platform displays the query text, the reference answer, the assigned task category (e.g., Reasoning QA, Trading \& Market), and the computed difficulty level. Alongside these metadata, the platform presents the candidate trajectories generated by the three frontier LLMs (Claude, Gemini, and GPT) in a side-by-side tabbed view. Each trajectory panel shows the full multi-turn interaction, including reasoning steps, tool invocations with their arguments, tool responses, and the final answer, along with summary statistics such as the number of turns and unique tools used.
Four financial domain experts used this platform to carry out two rounds of annotation. In the first round, experts independently reviewed a random sample of 100 queries to validate the LLM-as-Judge rankings. For each query, annotators selected the trajectory they judged to be the best based on correctness, tool usage efficiency, logical coherence, and answer completeness. We measured inter-annotator agreement and agreement with the automated rankings using Cohen's kappa, obtaining $\kappa$ = 0.89, which indicates strong alignment between human and model judgments. In the second round, experts reviewed the full set of 800 model-selected golden-label trajectories. When a trajectory contained errors—such as incorrect tool arguments, missing reasoning steps, or factually inaccurate final answers—the annotator flagged the specific issue and provided corrective feedback through the platform's annotation interface. The flagged trajectories were then revised by the language model based on this feedback, and the updated versions were re-reviewed by the experts until they met the required quality standard. This iterative human-in-the-loop process ensured that the final 800 golden-label trajectories are both reliable and representative of high-quality financial tool-calling behavior.

\clearpage
\section{The Details of Rubric Metric Definitions}
\label{app:rubric_details}

This appendix provides the formal definitions of all nine metrics in the rubric
evaluation protocol introduced in Section~\ref{sec:rubric}.

\paragraph{Action Correctness.}
This axis evaluates whether the agent selects and invokes the appropriate tools at
each step. \textbf{Tool-Calling F1}~\citep{chen2024t} computes a set-based F1 score
by comparing the tools invoked by the candidate trajectory against those in the
golden-label trajectory. To account for repeated invocations, we additionally compute
a bag-based F1 over the full tool-call sequence, capturing both precision and recall
at the individual call level. \textbf{Task Relevance} is an LLM-judged metric that
assesses whether each tool call is relevant to the given query, penalizing off-topic
or unnecessary invocations that may not be captured by F1 alone.

\paragraph{Execution Efficiency.}
This axis measures how concisely the agent completes the task. \textbf{Step Efficiency}
is defined as $\min\!\bigl(\frac{T_{\text{golden}}}{T_{\text{candidate}}},\,1.0\bigr)$,
where $T_{\text{golden}}$ and $T_{\text{candidate}}$ denote the number of turns in the
golden-label and candidate trajectories, respectively; a score of $1.0$ indicates that
the candidate completes the task in no more steps than the reference.
\textbf{Redundancy Score} is computed as
$1.0 - \frac{N_{\text{dup}}}{N_{\text{total}}}$, where $N_{\text{dup}}$ is the number
of exact-duplicate tool calls (identical tool name and arguments) and $N_{\text{total}}$
is the total number of tool calls. Together, these two metrics quantify whether the
agent reaches the correct answer without unnecessary detours or repeated operations.

\paragraph{Process Quality.}
This axis captures the qualitative characteristics of the reasoning trajectory beyond
tool selection and efficiency, and comprises three LLM-judged metrics.
\textbf{Logical Progression} evaluates whether the sequence of tool calls follows a
coherent and logical order, comparing the candidate against the golden-label trajectory
to identify misordered or illogical steps.
\textbf{Information Utilization} assesses whether the agent effectively incorporates
tool-call outputs into subsequent reasoning and the final answer, rather than ignoring
or misinterpreting retrieved information.
\textbf{Progress Score} measures whether each turn makes meaningful progress toward
answering the query, penalizing stalled, circular, or stagnant behavior.

\paragraph{Output Quality.}
This axis evaluates the end result of the agent's trajectory. \textbf{Task Pass Rate}
measures the factual correctness of the final answer against the reference answer.
\textbf{Final Answer Quality} provides a holistic assessment along the dimensions of
accuracy, completeness, and clarity, comparing the candidate's output against both the
reference answer and the golden-label trajectory.

\section{Prompts}

\subsection{Evaluation Rubric Prompts}
\label{appendix:rubric_prompts}
 
We present the LLM-judge prompts used in our rubric-based evaluation protocol. Our evaluation covers four axes: action correctness, execution efficiency, process quality, and output quality. The execution efficiency axis is computed algorithmically without LLM judges. The remaining three axes use LLM-judged prompts with a common scoring rubric.
 
\begin{tcolorbox}[
  colback=gray!8,
  colframe=gray!60,
  title={\textbf{Action Correctness}},
  fonttitle=\bfseries,
  breakable
]
 
\textbf{Pass Rate.}
You are an expert financial analyst evaluating the correctness of an AI assistant's final answer.
 
You will be given a financial question, a reference (ground truth) answer, and the candidate's final answer.
 
Evaluate whether the candidate's final answer is factually correct compared to the reference answer. Focus on key facts, figures, and data points. Minor formatting differences are acceptable; focus on substantive accuracy.
Scoring rubric (1-5):
\begin{itemize}[itemsep=3pt,parsep=1pt,leftmargin=1.5em]
    \item[\textbf{1}] - Completely wrong or no answer provided
    \item[\textbf{2}] - Major errors; only tangentially related to the correct answer
    \item[\textbf{3}] - Partially correct; some key facts match but significant information is wrong or missing
    \item[\textbf{4}] - Mostly correct; minor numerical or factual errors only
    \item[\textbf{5}] - Fully correct; matches reference answer in all key facts and figures
\end{itemize}
 
\end{tcolorbox}
 
\begin{tcolorbox}[
  colback=gray!8,
  colframe=gray!60,
  title={\textbf{Process Quality}},
  fonttitle=\bfseries,
  breakable
]
 
\textbf{Task Relevance.}
You are an expert evaluator of AI assistant tool-calling trajectories in the financial domain. You will evaluate whether each tool call in the candidate trajectory is relevant to answering the original query. A relevant tool call is one that could logically contribute information needed to answer the question.
Scoring rubric (1-5):
\begin{itemize}[itemsep=3pt,parsep=1pt,leftmargin=1.5em]
    \item[\textbf{1}] - Tool calls entirely unrelated to the query
    \item[\textbf{2}] - Mostly irrelevant tools called; fundamental misunderstanding of what data is needed
    \item[\textbf{3}] - Some relevant tools, but also many irrelevant or misguided calls
    \item[\textbf{4}] - Most tools are relevant; only minor irrelevant calls
    \item[\textbf{5}] - All tool calls are directly relevant to answering the query
\end{itemize}
 
\noindent\rule{\textwidth}{0.4pt}
 
\textbf{Logical Progression.}
You are an expert evaluator of AI assistant tool-calling trajectories in the financial domain. You will evaluate whether the sequence of tool calls follows a coherent, logical strategy. Each step should build on information from previous steps, and the overall sequence should represent a reasonable approach to answering the query. You will see both a golden (reference) trajectory showing one effective approach, and the candidate trajectory to evaluate. The candidate does NOT need to follow the same path as the golden trajectory --- it just needs to follow a logical strategy.
Scoring rubric (1-5):
\begin{itemize}[itemsep=3pt,parsep=1pt,leftmargin=1.5em]
    \item[\textbf{1}] - Chaotic; no coherent strategy; random or contradictory tool calls
    \item[\textbf{2}] - Weak logical flow; major gaps or illogical sequencing
    \item[\textbf{3}] - Some logical structure but with unnecessary detours or missed dependencies
    \item[\textbf{4}] - Clear logical flow with only minor inefficiencies
    \item[\textbf{5}] - Optimal logical progression; each step naturally follows from the previous
\end{itemize}
 
\noindent\rule{\textwidth}{0.4pt}
 
\textbf{Information Utilization.}
You are an expert evaluator of AI assistant tool-calling trajectories in the financial domain. You will evaluate how effectively the candidate uses the data returned by tool calls. Check whether key data from tool outputs is reflected in subsequent reasoning and the final answer, and whether any important information was ignored or misinterpreted.
Scoring rubric (1-5):
\begin{itemize}[itemsep=3pt,parsep=1pt,leftmargin=1.5em]
    \item[\textbf{1}] - Ignores tool outputs entirely; final answer does not reference retrieved data
    \item[\textbf{2}] - Uses tool outputs poorly; misinterprets data or draws wrong conclusions from it
    \item[\textbf{3}] - Uses some information but misses key data points available in tool outputs
    \item[\textbf{4}] - Good use of information; minor oversights in utilizing available data
    \item[\textbf{5}] - Excellent; fully leverages all relevant tool outputs in reasoning and final answer
\end{itemize}
 
\noindent\rule{\textwidth}{0.4pt}
 
\textbf{Progress Score.}
You are an expert evaluator of AI assistant tool-calling trajectories in the financial domain. You will evaluate whether each turn in the candidate trajectory makes meaningful progress toward answering the query. Look for signs of stalling, looping, repeated failed attempts, or wasted turns that do not advance the solution.
Scoring rubric (1-5):
\begin{itemize}[itemsep=3pt,parsep=1pt,leftmargin=1.5em]
    \item[\textbf{1}] - No progress made; stuck in loops or entirely off-track
    \item[\textbf{2}] - Minimal progress; many wasted turns with little forward movement
    \item[\textbf{3}] - Moderate progress; some turns are productive but others are wasted
    \item[\textbf{4}] - Good progress; nearly every turn advances toward the answer
    \item[\textbf{5}] - Every turn makes clear, meaningful progress toward the final answer
\end{itemize}
 
\end{tcolorbox}
 
\begin{tcolorbox}[
  colback=gray!8,
  colframe=gray!60,
  title={\textbf{Output Quality}},
  fonttitle=\bfseries,
  breakable
]
 
\textbf{Final Answer Quality.}
You are an expert financial analyst evaluating the overall quality of an AI assistant's final answer to a financial question.
 
Evaluate the answer considering accuracy (vs reference answer), completeness (addresses all parts of the question), clarity (well-organized and easy to understand), and presentation (appropriate use of formatting, tables, numbers).
 
You will also see the golden (reference) trajectory for context on what a high-quality response path looks like.
 Scoring rubric (1-5):
\begin{itemize}[itemsep=3pt,parsep=1pt,leftmargin=1.5em]
    \item[\textbf{1}] - No answer or completely unusable response
    \item[\textbf{2}] - Poor quality; major factual errors, missing most parts of the question, or incoherent
    \item[\textbf{3}] - Acceptable; addresses the question but incomplete, unclear, or has moderate errors
    \item[\textbf{4}] - Good quality; well-organized, mostly accurate, minor issues only
    \item[\textbf{5}] - Excellent; comprehensive, accurate, clearly presented, and addresses all parts
\end{itemize}
 
\end{tcolorbox}

\subsection{Evaluation System Prompt}
\label{appendix:system_prompts}

\begin{tcolorbox}[
  colback=gray!8,
  colframe=gray!60,
  title={\textbf{Evaluation Prompts}},
  fonttitle=\bfseries,
  breakable
]
 
\textbf{System Prompt.}
You are a financial assistant operating in an interactive environment with access to external tools.

Your goal is to answer user queries accurately and efficiently.

\textbf{User Prompt.} \{Query\}

\end{tcolorbox}

\clearpage

 
\section{Robustness Analysis Details}
\label{app:robustness}
 
This appendix reports the three studies summarized in
Section~\ref{ssec:robustness}: an out-of-family judge sensitivity study
(\S\ref{app:judge-sensitivity}), a human validation study of the
LLM-judged rubrics (\S\ref{app:human-validation}), and a sensitivity
analysis over metric aggregation schemes (\S\ref{app:aggregation}).
Table~\ref{tab:robustness-summary} collects the headline statistics.
 
\begin{table}[t]
\centering
\small
\begin{tabular}{llc}
\toprule
\textbf{Robustness check} & \textbf{Statistic} & \textbf{Value} \\
\midrule
\multirow{3}{*}{Out-of-family judge (DeepSeek-V4-Pro)}
  & Spearman's $\rho$ (overall) & 0.989 \\
  & Kendall's $\tau$ (overall)  & 0.949 \\
  & Pearson $r$ (overall)       & 0.994 \\
\midrule
\multirow{3}{*}{Human--human agreement (Krippendorff's $\alpha$)}
  & Task Relevance        & 0.709 \\
  & Logical Progression   & 0.979 \\
  & Final Answer Quality  & 0.567 \\
\midrule
\multirow{3}{*}{LLM--human consensus (weighted $\kappa_w$)}
  & Task Relevance        & 0.631 \\
  & Logical Progression   & 0.857 \\
  & Final Answer Quality  & 0.720 \\
\midrule
\multirow{3}{*}{Aggregation sensitivity (Kendall's $\tau$, worst case)}
  & Leave-one-metric-out  & 0.974 \\
  & Axis-balanced (2-/3-axis) & 0.974 \\
  & Median aggregation    & 0.974 \\
\bottomrule
\end{tabular}
\caption{Summary of the three robustness studies. Rank correlations are
computed over all 13 evaluated models; agreement coefficients are
computed over 90 human-annotated trajectories.}
\label{tab:robustness-summary}
\end{table}
 
\subsection{Out-of-Family Judge Sensitivity}
\label{app:judge-sensitivity}
 
\paragraph{Motivation and setup.}
In the main experiments, all LLM-judged metrics are scored by
Claude-Sonnet-4.6, while Tool-Calling F1, Step Efficiency, and Redundancy
Score are computed deterministically from the trajectory and are
therefore invariant to the choice of judge. Because Claude-family models
occupy the top of the leaderboard, a natural concern is judge-family
bias. To test this, we re-ran the full evaluation of all 13 models using
DeepSeek-V4-Pro as an alternative judge. DeepSeek-V4-Pro satisfies two
useful properties: it is not itself among the evaluated systems, and it
belongs to a different family from the GPT, Claude, and Gemini models
that dominate the top ranks. All prompts, rubrics, trajectories, and
decoding settings were held fixed; only the judge model was changed.
 
\paragraph{Results.}
Table~\ref{tab:judge-sensitivity} reports the six LLM-judged metrics and
the recomputed overall score under DeepSeek-V4-Pro. The three algorithmic
metrics are unchanged from Table~\ref{tab:main} by construction and are
omitted here. Claude-Opus-4.6 remains the top-ranked model. Only two
adjacent swaps occur relative to the published ranking: GPT-5.4 and
Claude-Sonnet-4.6 exchange ranks 2 and 3, and GPT-5-mini and
Deepseek-V3.2 exchange ranks 9 and 10. Across all 13 models the two
rankings agree closely: Spearman's $\rho = 0.989$, Kendall's
$\tau = 0.949$, and Pearson $r = 0.994$ between overall scores.
 
\begin{table}[t]
\centering
\small
\begin{tabular}{clcccccc c}
\toprule
\textbf{Rank} & \textbf{Model} & \textbf{Pass} & \textbf{Task} &
\textbf{Logical} & \textbf{Info} & \textbf{Progress} &
\textbf{Answer} & \textbf{Overall} \\
 & & \textbf{Rate} & \textbf{Relev.} & \textbf{Prog.} &
\textbf{Util.} & \textbf{Score} & \textbf{Quality} & \\
\midrule
1  & Claude-Opus-4.6   & 1.93 & 4.29 & 4.28 & 2.84 & 4.11 & 2.93 & 0.767 \\
2  & GPT-5.4           & 2.60 & 4.04 & 3.72 & 2.70 & 3.44 & 2.79 & 0.734 \\
3  & Claude-Sonnet-4.6 & 1.94 & 4.28 & 3.72 & 2.77 & 4.04 & 2.76 & 0.731 \\
4  & Gemini-3.1-Pro    & 1.92 & 3.95 & 3.32 & 2.19 & 3.26 & 2.43 & 0.666 \\
5  & KiMi-2.5          & 1.71 & 3.82 & 3.18 & 1.95 & 3.27 & 2.39 & 0.643 \\
6  & Qwen3.5-397B      & 2.00 & 3.84 & 2.92 & 2.21 & 3.23 & 2.51 & 0.642 \\
7  & Qwen3.5-27B       & 1.76 & 4.12 & 2.94 & 1.93 & 3.11 & 2.10 & 0.621 \\
8  & Gemini-3-Flash    & 1.85 & 3.82 & 2.37 & 2.10 & 2.82 & 2.31 & 0.610 \\
9  & GPT-5-mini        & 1.51 & 3.69 & 2.90 & 1.63 & 3.07 & 1.96 & 0.596 \\
10 & Deepseek-V3.2     & 1.46 & 3.98 & 2.78 & 1.70 & 3.01 & 2.30 & 0.596 \\
11 & OSS-120B          & 1.49 & 3.69 & 2.55 & 1.74 & 2.74 & 1.82 & 0.559 \\
12 & Llama-3.3-70B     & 1.82 & 3.71 & 1.51 & 2.21 & 1.88 & 1.85 & 0.544 \\
13 & Qwen-3.5-9B       & 1.49 & 1.71 & 2.06 & 1.08 & 1.72 & 1.55 & 0.457 \\
\bottomrule
\end{tabular}
\caption{Evaluation under the out-of-family judge DeepSeek-V4-Pro. All
LLM-judged metrics are on a $[1,5]$ scale. Tool-Calling F1, Step
Efficiency, and Redundancy Score are computed deterministically and are
identical to Table~\ref{tab:main}; the Overall score is the mean of all
nine metrics after normalization to $[0,1]$.}
\label{tab:judge-sensitivity}
\vspace{-10pt}
\end{table}
 
\paragraph{No self-enhancement bias.}
A judge might be expected to favor models from its own family. We find no
such effect: under the DeepSeek-V4-Pro judge, Deepseek-V3.2 remains in
the lower-middle group at rank 10 with an overall score of $0.596$,
rather than being promoted above the GPT, Claude, Gemini, Kimi, or
stronger Qwen systems. Its relative position is unchanged from the
Claude-judged leaderboard.
 
\paragraph{Where the two judges differ.}
The judges do differ in absolute calibration. DeepSeek-V4-Pro is
noticeably stricter on Pass Rate (e.g., $2.65 \rightarrow 1.93$ for
Claude-Opus-4.6) and more generous on Progress Score and Task Relevance.
This is expected: Likert-scale judges anchor differently even under
identical rubrics. What matters for a benchmark is that the induced
ordering is stable, which the rank correlations confirm. It also
reinforces a point we make in Section~\ref{ssec:robustness}: FinTrace
scores should be read comparatively across models under a fixed judge,
not as calibrated absolute quantities. For this reason we release the
judge configuration, prompts, and raw per-query scores so that future
work can either reuse our judge or re-score the released trajectories
under its own.

\subsection{Human Validation of the LLM-Judged Rubrics}
\label{app:human-validation}
 
\paragraph{Study design.}
The nine rubrics in FinTrace were originally drafted and refined with
input from financial domain experts, so that the criteria are both
financially meaningful and directly comparable to human assessment. To
quantify how closely the LLM judge reproduces human scoring, we ran a
dedicated annotation study on the three LLM-judged metrics most exposed
to inter-annotator disagreement: Task Relevance, Logical Progression, and
Final Answer Quality (all on a 1--5 Likert scale). We selected three
models spanning the performance spectrum, GPT-5.4 (frontier),
Qwen3.5-397B (strong open-weight), and Deepseek-V3.2 (mid-tier), and
randomly sampled 10 trajectories per model per metric, giving 30
annotated items per model and 90 in total. Each item was scored
independently by two annotators, with a distinct pair assigned to each
model (six annotators in three pairs). Annotators were blinded to model
identity to prevent prior-driven bias, and used the same annotation
platform described in Appendix~\ref{app:annotation} with the rubric text
of Appendix~\ref{appendix:rubric_prompts} shown inline.
 
\paragraph{Result 1: the rubric separates models under human scoring.}
Table~\ref{tab:human-scores} reports per-annotator scores and the human
consensus (the mean of the two annotators for each model). The consensus
ranks the three models in the expected order, GPT-5.4 $>$ Qwen3.5-397B
$>$ Deepseek-V3.2, on all three metrics, with clear separation in the
average score ($3.62$ vs.\ $3.12$ vs.\ $2.51$). Critically, this ordering
holds for \emph{each individual annotator} as well, not only for the
consensus. Even where raters apply different absolute scales, the
relative ordering induced by the rubric is stable, which is the property
a comparative benchmark requires.

\begin{table}[t]
\centering
\footnotesize
\setlength{\tabcolsep}{3pt}
\renewcommand{\arraystretch}{1}

\begin{tabular}{llcccc}
\toprule
\textbf{Model}
& \textbf{Annotator}
& \makecell{\textbf{Final Answer}\\\textbf{Quality}}
& \makecell{\textbf{Task}\\\textbf{Relevance}}
& \makecell{\textbf{Logical}\\\textbf{Progression}}
& \textbf{Average} \\
\midrule

\multirow{3}{*}{GPT-5.4}
& A         & 3.20 & 4.20 & 3.70 & 3.70 \\
& B         & 2.40 & 4.30 & 3.90 & 3.53 \\
& Consensus & 2.80 & 4.25 & 3.80 & \textbf{3.62} \\
\midrule

\multirow{3}{*}{Qwen3.5-397B}
& C         & 2.90 & 3.80 & 3.20 & 3.30 \\
& D         & 2.10 & 3.70 & 3.00 & 2.93 \\
& Consensus & 2.50 & 3.75 & 3.10 & \textbf{3.12} \\
\midrule

\multirow{3}{*}{DeepSeek-V3.2}
& E         & 1.90 & 3.10 & 2.80 & 2.60 \\
& F         & 1.70 & 2.95 & 2.60 & 2.42 \\
& Consensus & 1.80 & 3.03 & 2.70 & \textbf{2.51} \\
\bottomrule
\end{tabular}

\caption{Human annotation scores on 90 sampled trajectories. Consensus is
the mean of the two annotators assigned to each model. The model ordering
is identical for every annotator and for the consensus across all three
metrics.}
\label{tab:human-scores}
\end{table}
 
\paragraph{Result 2: annotators interpret the process rubrics consistently.}
Table~\ref{tab:agreement} reports Krippendorff's $\alpha$ between the two
annotators of each pair. Logical Progression ($\alpha = 0.979$) and Task
Relevance ($\alpha = 0.709$) reach high to very high agreement,
indicating that these rubric definitions are read consistently across
raters. Final Answer Quality shows lower agreement ($\alpha = 0.567$).
Inspecting the disagreements, this reflects legitimate differences in how
raters weigh completeness, clarity, and numerical accuracy when
collapsing a holistic judgment into a single 1--5 score, rather than
disagreement about which model is better: as Table~\ref{tab:human-scores}
shows, both annotators produce the same model ordering on Final Answer
Quality despite differing by up to $0.8$ points in absolute level.
 
\begin{table}[t]
\centering
\small
\begin{tabular}{lcc}
\toprule
\textbf{Metric} & \textbf{$\alpha$ (Human--Human)} &
\textbf{$\kappa_w$ (LLM--Human Consensus)} \\
\midrule
Task Relevance       & 0.709 & 0.631 \\
Logical Progression  & 0.979 & 0.857 \\
Final Answer Quality & 0.567 & 0.720 \\
\bottomrule
\end{tabular}
\caption{Inter-annotator agreement (Krippendorff's $\alpha$) and
agreement between the LLM judge and the human consensus (quadratic
weighted Cohen's $\kappa$).}
\label{tab:agreement}
\vspace{-10pt}
\end{table}
 
\paragraph{Result 3: the LLM judge aligns with human consensus.}
The LLM judge reaches substantial agreement with the human consensus on
all three metrics ($\kappa_w = 0.631$--$0.857$). On Final Answer Quality,
LLM--human agreement ($\kappa_w = 0.720$) exceeds human--human agreement
($\alpha = 0.567$). This is not contradictory. Because the judge is
compared against the \emph{averaged} human score, it is insulated from
the individual calibration offsets that drive the human--human
disagreement; in effect the judge behaves as a central-tendency
evaluator, producing scores that fall between those of the two individual
raters. This mirrors observations in prior LLM-as-a-judge work
\citep{zheng2023judging}.
 
\paragraph{Takeaways and limitations.}
Together these results support the validity of the rubric-based protocol:
(i) the rubric separates models in the expected order under human
scoring, (ii) the process-quality rubrics are interpreted consistently
across annotators, and (iii) the LLM judge reaches
substantial-to-high agreement with the human consensus. We are explicit
about the residual limitation: Final Answer Quality is most reliable as a
\emph{relative} ranking signal rather than as a calibrated absolute
score, and readers should compare models on it rather than interpret the
raw value. We also note the scope of this study---90 trajectories, three
metrics, three models---which was chosen to keep double annotation
feasible under expert-annotator cost. Extending human validation to all
six LLM-judged metrics and to the full model set is left to future
releases of the benchmark.

\subsection{Sensitivity of the Ranking to Metric Aggregation}
\label{app:aggregation}
 
The Overall score reported in Table~\ref{tab:main} is the unweighted mean
of the nine normalized metrics. Under this scheme, the six LLM-judged
metrics implicitly carry $6/9 \approx 67\%$ of the weight against
$3/9 \approx 33\%$ for the three algorithmic metrics. We therefore test
whether the leaderboard is an artifact of this particular aggregation
using three complementary analyses. Across all of them, the published
ranking is preserved with Kendall's $\tau \geq 0.974$ and an invariant
top-3 set.
 
\paragraph{(1) Leave-one-metric-out (LOMO).}
For each metric $m$, we recompute the Overall score as the mean of the
remaining eight normalized metrics, re-rank all 13 models, and compare
against the published ranking. Results are in Table~\ref{tab:lomo}. No
single metric drives the ranking: the worst-case Kendall's $\tau$ across
all nine drops is $0.974$, well above the conventional $\tau = 0.8$
strong-agreement threshold, and five of the nine drops reproduce the
published ranking exactly. The top-3 set
$\{$Claude-Opus-4.6, Claude-Sonnet-4.6, GPT-5.4$\}$ is invariant under
every drop.
 
\begin{table}[t]
\centering
\footnotesize
\begin{tabular}{lccc}
\toprule
\textbf{Dropped metric} & \textbf{Kendall's $\tau$} &
\textbf{Top-3 overlap} & \textbf{Top-5 overlap} \\
\midrule
Tool-Calling F1          & 0.974 & 1.00 & 1.00 \\
Step Efficiency          & 0.974 & 1.00 & 0.80 \\
Redundancy Score         & 1.000 & 1.00 & 1.00 \\
Pass Rate                & 1.000 & 1.00 & 1.00 \\
Task Relevance           & 1.000 & 1.00 & 1.00 \\
Logical Progression      & 1.000 & 1.00 & 1.00 \\
Information Utilization  & 0.974 & 1.00 & 1.00 \\
Progress Score           & 0.974 & 1.00 & 1.00 \\
Final Answer Quality     & 1.000 & 1.00 & 1.00 \\
\midrule
\textbf{Worst case}      & \textbf{0.974} & \textbf{1.00} & \textbf{0.80} \\
\bottomrule
\end{tabular}
\caption{Leave-one-metric-out sensitivity. Rank correlations are computed
against the published ranking over all 13 models.}
\label{tab:lomo}
\end{table}
 
\paragraph{(2) Axis-balanced aggregation.}
We next re-balance the two metric families so that the algorithmic and
LLM-judged groups receive equal weight. The 2-axis variant averages the
two family means,
\begin{equation}
\text{Overall}_{\text{2-axis}}
= \tfrac{1}{2}\,\overline{\text{algorithmic}}
+ \tfrac{1}{2}\,\overline{\text{LLM-judged}},
\end{equation}
which raises each algorithmic metric's effective weight from $11.1\%$ to
$16.7\%$ ($\times 1.5$) and lowers each LLM-judged metric's weight to
$8.3\%$ ($\times 0.75$). The 3-axis variant further splits the LLM-judged
family into a process group (task relevance, logical progression,
information utilization, progress score) and an outcome group (pass rate,
final answer quality),
\begin{equation}
\text{Overall}_{\text{3-axis}}
= \tfrac{1}{3}\,\overline{\text{tool-use}}
+ \tfrac{1}{3}\,\overline{\text{process}}
+ \tfrac{1}{3}\,\overline{\text{outcome}}.
\end{equation}
As a worked example, for Claude-Opus-4.6 the algorithmic mean is
$(0.896 + 0.926 + 0.997)/3 = 0.940$ and the LLM-judged mean is
$(0.530 + 0.828 + 0.902 + 0.646 + 0.698 + 0.668)/6 = 0.712$, giving
$\text{Overall}_{\text{2-axis}} = 0.826$. Both variants yield the same
ranking, which differs from the published one by exactly one mid-table
swap (Deepseek-V3.2 $\leftrightarrow$ Gemini-3-Flash at ranks 8/9); the
top seven positions are identical (Table~\ref{tab:aggregation}). The
leaderboard is therefore not an artifact of LLM-judged metrics being
implicitly over-weighted.
 
\paragraph{(3) Median aggregation.}
The arithmetic mean can be lifted by one or two unusually strong metrics
even when most metrics are mediocre; the median is determined by the
central value alone and is insensitive to extremes at either end. We
replace the mean with the median of the nine normalized values. For
Claude-Opus-4.6 the sorted values are
$0.530,\ 0.646,\ 0.668,\ 0.698,\ \mathbf{0.828},\ 0.896,\ 0.902,\ 0.926,\ 0.997$,
so the median is $0.828$ against a mean of $0.788$. As an intuition for
the robustness this buys: if Pass Rate fell from $0.530$ to $0.000$, the
mean would drop to $0.729$ while the median would stay at $0.828$. The
median ranking differs from the published one by the same single
mid-table swap as the axis-balanced variants, with the top seven
unchanged. No model therefore reaches a high overall rank by virtue of
outlier values on one or two metrics.
 
\begin{table}[t]
\centering
\footnotesize
\begin{tabular}{lccc}
\toprule
\textbf{Aggregation variant} & \textbf{Kendall's $\tau$} &
\textbf{Top-3 overlap} & \textbf{Top-5 overlap} \\
\midrule
2-axis (algorithmic / LLM-judged)      & 0.974 & 1.00 & 1.00 \\
3-axis (tool-use / process / outcome)  & 0.974 & 1.00 & 1.00 \\
Median aggregation                     & 0.974 & 1.00 & 1.00 \\
\bottomrule
\end{tabular}
\caption{Alternative aggregation schemes compared against the published
ranking. All three preserve the top-7 ordering and differ only by one
mid-table swap.}
\label{tab:aggregation}
\end{table}

\clearpage
\section{Trajectory Cleaning Details for Post-training}
\label{appendix:cleaning}

\subsection{Cleaning Transformations}

Raw trajectories contain artifacts from the data collection process that must be removed before training. We apply three cleaning transformations to each conversation:

\begin{tcolorbox}[
  colback=gray!8,
  colframe=gray!60,
  title={\textbf{System Prompt For Training}},
  fonttitle=\bfseries,
  breakable
]
 
\textbf{System Prompt.}
You are a financial assistant operating in an interactive environment with access to external tools.

Your goal is to answer user queries accurately and efficiently.

\end{tcolorbox}

\begin{itemize}
    \item \textbf{System prompt replacement.} The original system prompt contains data-source-specific tool-calling protocol instructions. We replace it with a generic assistant prompt so the model does not learn a protocol that will not exist at inference time.
    
    \item \textbf{Ghost tool-call removal.} During data collection, the source model occasionally produced tool calls in XML format that were captured as plain text but never executed---no tool response follows them. We detect these by identifying assistant messages that (a) lack a \texttt{tool\_calls} field, (b) contain malformed XML patterns such as \texttt{<function\_calls>} or \texttt{<function=}, and (c) are not followed by a tool response.
    
    \item \textbf{Reasoning relocation.} When an assistant message contains both reasoning text (in \texttt{content}) and tool calls (in \texttt{tool\_calls}), we move the reasoning to the \texttt{reasoning\_content} field. The target model's chat template renders this field inside \texttt{<think>} tags, ensuring that tool-call reasoning is properly formatted during training and inference.
\end{itemize}

\subsection{Filtering Details}
After cleaning, we apply the following filters to remove low-quality examples:

\begin{itemize}
    \item \textbf{All-empty filter.} Removes examples where \emph{every} tool response is effectively empty (empty string, \texttt{[]}, or \texttt{\{\}}). Importantly, we retain examples with \emph{partial} empty responses: manual inspection of these records reveals that most of them demonstrate successful failure-recovery behavior, where the model received an empty result, retried with different tools or parameters, and eventually retrieved useful data. These recovery patterns provide valuable training signal.

    \item \textbf{Token-length filter.} We tokenize each example using the target model's chat template (including tool schemas) and remove examples exceeding 16,384 (16k).
\end{itemize}

\subsection{Tool Augmentation Details}
\begin{itemize}
    \item \textbf{Called tools.} are always included in the pool.
    \item \textbf{Similar tools.} fill 50\% of the remaining slots. We embed each tool as a text string of the form ``\texttt{\{name\}: \{description\}}'' using Voyage Finance 2, a finance-domain embedding model. For each candidate tool in the pool, we compute its cosine similarity to every called tool and take the maximum across called tools. The candidates with the highest max-similarity scores are selected.
    \item \textbf{Random tools.} fill the remaining slots by uniform sampling from all tools not yet selected.
\end{itemize}

\subsection{Preference Pair Construction Details}
To construct the preference dataset for DPO training, we follow a three-stage procedure:

\begin{itemize}
    \item \textbf{Rollout.} We use the best SFT checkpoint to generate trajectories on the queries in the held-out DPO split. Each rollout executes tool calls against the live FMP MCP server, producing realistic multi-turn trajectories.

    \item \textbf{LLM Judge.} We use Claude Sonnet 4.6 as a pairwise judge to compare each model-generated (rollout) trajectory against the corresponding reference trajectory. To mitigate position bias, we randomize which trajectory appears as ``Response A'' and which as ``Response B'' in each comparison. The judge outputs one of three verdicts: chosen is better, rejected is better, or tie.

    \item \textbf{Filtering.} We retain only pairs where the reference trajectory is judged to be better than the model-generated trajectory (\emph{chosen\_is\_better}), discarding ties and cases where the model-generated trajectory wins. This filtering ensures a clean preference signal for DPO training.
\end{itemize}

\begin{tcolorbox}[
  colback=gray!8,
  colframe=gray!60,
  title={\textbf{LLM as Judge Prompt for Trajectory Comparison}},
  fonttitle=\bfseries,
  breakable
]

\textbf{LLM as Judge Prompt.}
You are an expert evaluator of AI financial assistants. You will compare two responses to the same user query. Both responses use tool calls to fetch financial data before answering.

\vspace{4pt}
\textbf{Evaluate based on these criteria:}
\begin{itemize}[leftmargin=1.5em, itemsep=2pt]
    \item \textbf{Correctness} — Is the final answer factually accurate given the tool results?
    \item \textbf{Tool Usage} — Did the assistant call the right tools with correct parameters? Were calls efficient?
    \item \textbf{Completeness} — Does the answer address all parts of the user's question?
    \item \textbf{Coherence} — Is the reasoning clear and the answer well-structured?
\end{itemize}

\vspace{4pt}
\textbf{User Query:} \texttt{\{user\_query\}}

\vspace{2pt}
\textbf{Available Tools:} \texttt{\{tool\_summary\}}

\vspace{2pt}
\textbf{Response A:} \texttt{\{trajectory\_a\}}

\vspace{2pt}
\textbf{Response B:} \texttt{\{trajectory\_b\}}

\vspace{6pt}
\textbf{Forced tool-call schema for structured output:}\\
Verdict options: \texttt{"A is better"}, \texttt{"B is better"}, \texttt{"same quality"}\\
Required fields: \texttt{verdict} (one of the three options) + \texttt{reasoning} (brief 2--3 sentence explanation)

\end{tcolorbox}

\clearpage
\section{Training Details}
\label{appendix:training}

\subsection{Base Models}
We fine-tune Qwen 3 8B and Qwen 3 32B models. We use LoRA~\citep{hu2021loralowrankadaptationlarge} for parameter-efficient fine-tuning and are trained on 8$\times$ NVIDIA H200 NVL GPUs (143\,GB each) using FSDP2 with full sharding and bf16 mixed precision.

\subsection{Qwen 3 8B}
 
\begin{table}[h]
\centering
\label{tab:sft_config}
\begin{tabular}{@{}lcc@{}}
\toprule
\textbf{Parameter} & \textbf{8B} \\
\midrule
Learning rate & $1 \times 10^{-5}$ \\
Epochs & 15.9 \\
Effective batch size & 24  \\
Max sequence length & 16,384 \\
LoRA rank ($r$) & 128  \\
LoRA alpha ($\alpha$) & 128 \\
LoRA dropout & 0.0  \\
LoRA targets & all-linear  \\
\bottomrule
\end{tabular}
\caption{Training configuration for Qwen 3 8B.}
\end{table}

\begin{table}[h]
\centering
\label{tab:dpo_config}
\begin{tabular}{@{}lcc@{}}
\toprule
\textbf{Parameter} & \textbf{8B} \\
\midrule
Learning rate & $1 \times 10^{-6}$ \\
Epochs & 5.7 \\
Effective batch size & 32 \\
Max sequence length & 16,384 \\
LoRA rank ($r$) & 128  \\
LoRA alpha ($\alpha$) & 128  \\
LoRA dropout & 0.0  \\
LoRA targets & all-linear  \\
$\beta$ (KL penalty) & 0.05 \\
\bottomrule
\end{tabular}
\caption{DPO training configuration for Qwen 3 8B.}
\end{table}

\begin{table}[h]
\centering
\label{tab:rollout_config}
\begin{tabular}{@{}lc@{}}
\toprule
\textbf{Parameter} & \textbf{Value} \\
\midrule
Max sequence length & 28,000 \\
Max turns per rollout & 7 \\
Sampling (top-$p$) & 0.95 \\
\bottomrule
\end{tabular}
\caption{Rollout configuration for rejected trajectory generation for Qwen 3 8B.}
\end{table}

\begin{table}[h!]
\centering
\label{tab:rollout_config2}
\begin{tabular}{@{}lc@{}}
\toprule
\textbf{Parameter} & \textbf{Value} \\
\midrule
Max sequence length & 28,000 \\
Max turns per rollout & 7 \\
Sampling (top-$p$) & 0.95 \\
\bottomrule
\end{tabular}
\caption{Rollout configuration for trajectory evaluation for Qwen 3 8B.}
\end{table}

\subsection{Qwen 3 32B}
\vspace{-10pt}

\begin{table}[h]
\centering
\label{tab:sft_config_32b}
\begin{tabular}{@{}lcc@{}}
\toprule
\textbf{Parameter} & \textbf{32B} \\
\midrule
Learning rate & $5 \times 10^{-5}$ \\
Epochs & 7.2 \\
Effective batch size & 64  \\
Max sequence length & 16,384 \\
LoRA rank ($r$) & 128  \\
LoRA alpha ($\alpha$) & 128 \\
LoRA dropout & 0.0  \\
LoRA targets & all-linear  \\
\bottomrule
\end{tabular}
\caption{Training configuration for Qwen 3 32B.}
\vspace{-10pt}
\end{table}

\begin{table}[h]
\centering
\label{tab:dpo_config2}
\begin{tabular}{@{}lcc@{}}
\toprule
\textbf{Parameter} & \textbf{8B} \\
\midrule
Learning rate & $1 \times 10^{-6}$ \\
Epochs & 6 \\
Effective batch size & 32 \\
Max sequence length & 16,384 \\
LoRA rank ($r$) & 128  \\
LoRA alpha ($\alpha$) & 128  \\
LoRA dropout & 0.0  \\
LoRA targets & all-linear  \\
$\beta$ (KL penalty) & 0.05 \\
\bottomrule
\end{tabular}
\caption{DPO training configuration for Qwen 3 32B.}
\vspace{-10pt}
\end{table}

\begin{table}[h]
\centering
\label{tab:rollout_config}
\begin{tabular}{@{}lc@{}}
\toprule
\textbf{Parameter} & \textbf{Value} \\
\midrule
Max sequence length & 28,000 \\
Max turns per rollout & 7 \\
Sampling (top-$p$) & 0.95 \\
\bottomrule
\end{tabular}
\caption{Rollout configuration for rejected trajectory generation for Qwen 3 32B.}
\vspace{-20pt}
\end{table}

\begin{table}[h]
\centering
\label{tab:rollout_config}
\begin{tabular}{@{}lc@{}}
\toprule
\textbf{Parameter} & \textbf{Value} \\
\midrule
Max sequence length & 28,000 \\
Max turns per rollout & 7 \\
Sampling (top-$p$) & 0.95 \\
\bottomrule
\end{tabular}
\caption{Rollout configuration for trajectory evaluation for Qwen 3 32B.}
\end{table}

\end{document}